\documentclass{article}
\usepackage{jfrExamplee}
\usepackage{graphicx}
\usepackage{apalike}
\usepackage{setspace}
\usepackage{hyperref}
\usepackage{esvect}
\usepackage{siunitx}
\usepackage{amsfonts} 

\usepackage[utf8]{inputenc}

\usepackage[ruled,vlined,linesnumbered]{algorithm2e}

 \usepackage{array,multirow,graphicx}
\usepackage{subcaption}
\usepackage{amsmath}
\usepackage{cases}

\usepackage{makecell}

\usepackage{mathtools}
\DeclarePairedDelimiter\ceil{\lceil}{\rceil}

\let\given\givenbase

\usepackage[utf8]{inputenc} 
\usepackage[T1]{fontenc}

\usepackage[table]{xcolor}
\definecolor{Gray}{gray}{0.9}
\definecolor{LightCyan}{rgb}{0.88,1,1}

\usepackage{booktabs}

\newcommand{\chl}{\cellcolor{Gray}} 

\newcommand{\mathbigspace}{\;\;\;\;}

\numberwithin{equation}{section} 

\graphicspath{{figs/}}
\DeclareGraphicsExtensions{.pdf,.jpeg,.jpg,.png}

\title{UAV-Based Human Body Detector Selection and Fusion for Geolocated Saliency Map Generation}

\author{
Piotr Rudol$^1$, Patrick Doherty$^{1,2}$, Mariusz Wzorek$^1$, and Chattrakul Sombattheera$^2$  \\
$^1$Department of Computer and Information Science\\
Link{\"o}ping University\\
S-581 83 Link{\"o}ping, Sweden \\
\texttt{\{piotr.rudol|patrick.doherty|mariusz.wzorek\}@liu.se} \\
$^2$Faculty of Informatics\\
Mahasarakham Univeristy\\
\texttt{chattrakul.s@msu.ac.th} 
}

\begin{document}

\maketitle


\begin{abstract}
The problem of reliably detecting and geolocating objects of different classes in soft real-time is essential in many application areas, such as Search and Rescue performed using Unmanned Aerial Vehicles (UAVs). This research addresses the complementary problems of system contextual vision-based detector selection, allocation, and execution, in addition to the fusion of detection results from teams of UAVs for the purpose of accurately and reliably geolocating objects of interest in a timely manner. In an offline step, an application-independent evaluation of vision-based detectors from a system perspective is first performed. Based on this evaluation, the most appropriate algorithms for online object detection for each platform are selected automatically before a mission, taking into account a number of practical system considerations, such as the available communication links, video compression used, and the available computational resources. The detection results are fused using a method for building maps of salient locations which takes advantage of a novel sensor model for vision-based detections for both positive and negative observations. A number of simulated and real flight experiments are also presented, validating the proposed method.

\end{abstract}


\section{Introduction}\label{sec:intro}

The use of Unmanned Aerial Vehicles has been shifting in recent years from mainly military to civil sector applications, including the area of Search and Rescue (SAR). Many current systems can be operated with minimal training as remote cameras to obtain eye-in-the-sky views for use by emergency rescue and other services. Increasing the autonomy of such systems is the next step in the development process, both in terms of efficiency of operation as well as more sophisticated data analysis. 
The application area which is addressed in this paper requires a generic search for objects in outdoor environments using UAVs equipped with imaging sensors. An example instantiation of such a generic scenario is searching for missing persons or finding victims of natural or man-made disasters. In both cases, the goal is to obtain geographical locations of objects of interest, humans in this case, with minimal user involvement in mission preparation, execution, as well as data analysis. The output of such search and rescue missions is the generation of saliency maps that geolocate objects of interest. The resulting maps of salient objects can subsequently be used, for example, to deliver medical supplies or direct human rescuers to people in need. 

A substantial body of research and commercial systems have been developed to address various challenges associated with the above application, including state estimation, sensor fusion, control, path and mission planning, sensor utilization planning, flight coordination, and more. Simultaneously, sensing technologies and the corresponding algorithms have witnessed steady improvements. Vision-based algorithms have made the most significant advancements in recent years. This progress can be attributed to developments in the field of Big Data and Deep Learning, in addition to the utilization of specialized and highly powerful hardware, particularly Graphics Processing Units (GPUs). Over the past decade, the number of techniques, application areas, and the accuracy of results in image processing have experienced rapid growth. Deep Convolutional Neural Networks (DCNNs or CNNs), which are commonly used for image analysis, have significantly enhanced and improved various image processing tasks, such as classification, detection, and semantic segmentation.

\begin{figure}[t]
    \centering
    \includegraphics[width=1.0\columnwidth]{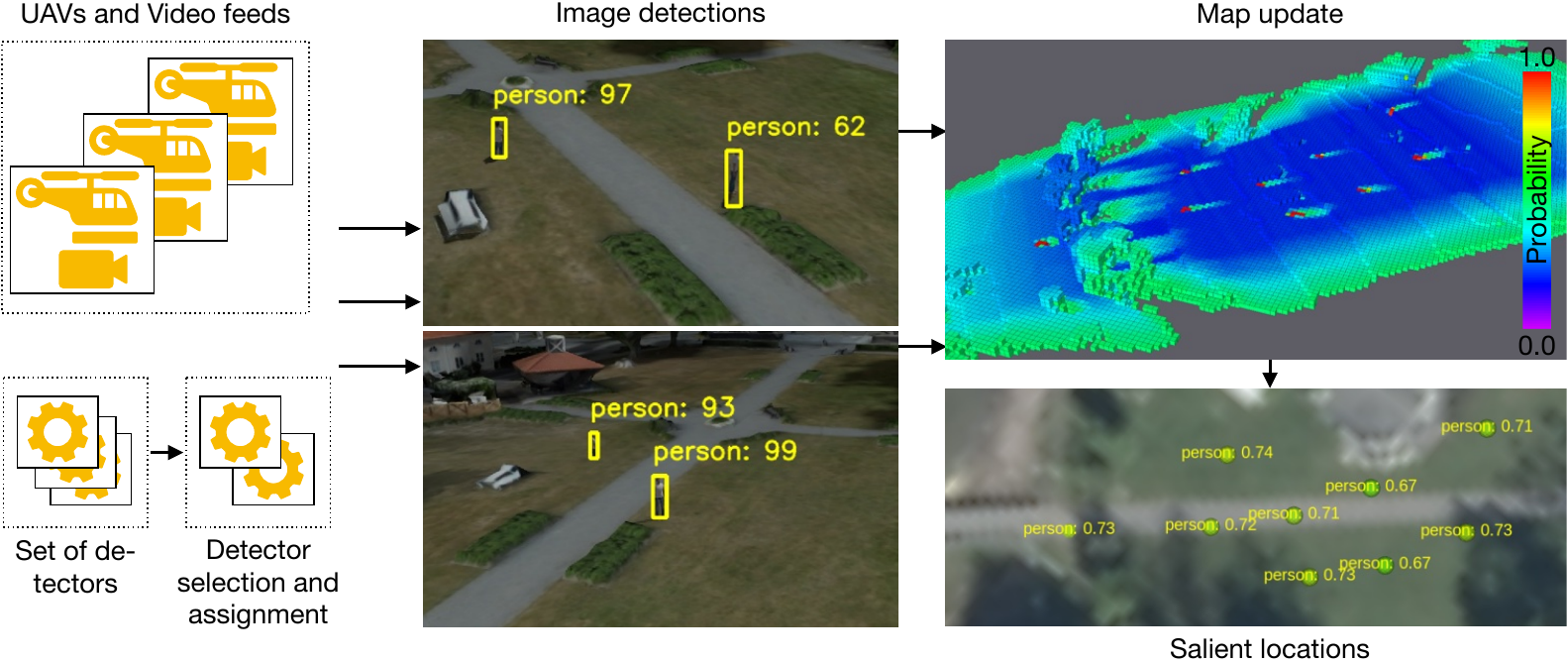}
    \caption{Overview of the main components of the proposed system. Video streams from one or more UAV platforms are processed using vision-based detection algorithms (selected optimally) to produce results in the form of bounding boxes and confidence scores. The detections are fused in the form of a map which represents probabilities of locations containing objects of specific classes. Based on this map, a list of salient locations is computed in the form of 3D locations. }
    \label{fig:overview_main}
\end{figure}

In this paper, these advances in image processing are leveraged to examine the broader issue of detecting and geolocating objects from a system perspective, considering not only the choice of the object detection algorithms used but also the practical constraints of operating in the field associated with the particular hardware, communication and computational configurations used by participating UAV platforms. An overview of the proposed approach is presented in Figure~\ref{fig:overview_main}.
First, an evaluation is made of a range of existing object detection neural networks based on their relative performance using various metrics specified later in the paper. Additionally,  the execution time properties of these networks are quantified relative to the use of different hardware and communication platforms intended to be used in an operational mission. This data enables one to optimize the specific allocation of the available computational resources among teams of UAVs and to execute the most relevant object detectors during an operational mission, taking into account physical limitations such as the available data link bandwidth in a specific operational environment. The evaluation and configuration are done prior to the execution of a specific mission, where an optimal configuration is instantiated for the team of UAVs intended to be used. During an ensuing mission, the collection of vision-based detections generated are then fused, using a novel probabilistic fusion algorithm proposed in the paper. Output from this algorithm is then used to generate a common geolocated map of the detected objects. This saliency map encodes the probabilities of the locations in the environment containing (or not containing) objects of different classes. The saliency map can then be used by human or robotic operators to define new missions associated with the newly detected geolocated objects.

\subsection{Background}

Designing a system to address the outlined task presents several challenges and often comes with limitations. The target application context is outdoor emergency SAR operations involving teams of diverse systems, including smaller quadrotor drone platforms working alongside human rescue responders and other resources. These teams collaborate to achieve complex goals, such as aerial-based human body detection and the timely delivery of medical and food supplies to potentially injured victims. Within this context, two main groups of issues need to be considered at the system level: vision-based object detection and the properties of the available hardware. These two groups influence each other in both directions. 

When it comes to vision-based object detection, the first issue is small \emph{object sizes}, which results from a combination of the flight altitude and camera parameters. A related issue is the \emph{object orientation}, which results from the flight direction and the camera gimbal orientation. Since the objects of interest, such as people, can behave freely in the environment, the \emph{object appearance} can vary substantially due to, for example, sitting, lying on the ground, or walking (\emph{mobility}). Due to the same problem, \emph{occlusion} can pose an issue. \emph{Observation time}, which is a function of the camera's field of view, as well as the speed and altitude of flight, must also be taken into account. 

Additionally, the properties of the available computational and communication hardware impose several challenges that are not often considered in this context.  For a particular UAV platform, there are options for executing computationally intensive image processing algorithms on-board the system itself; leveraging resources on another robotic system such as a UAV; outsourcing some or all of the computation to workstations in Ground Operations centers (GOPs); using Cloud resources for computation; or using different combinations of the above. This is an interesting optimization problem.

For many computer vision algorithms, it is assumed that the input images are of high fidelity. This is, however, not the case in application areas such as surveillance or, in fact, in robotics due to the way the acquisition, transmission, and even storage of video data is performed. This is further worsened in the case of aerial robots as the carrying capacity generally does not allow for the use of high-fidelity imaging sensors and sufficient computational power. Moreover, for very small-size UAV platforms, onboard image processing is severely limited as the latest developments and state-of-the-art algorithms using deep neural networks require computational power offered mainly by GPUs in order to deliver timely results. 

\begin{figure}[h]
   \centering
   \includegraphics[width=0.95\columnwidth]{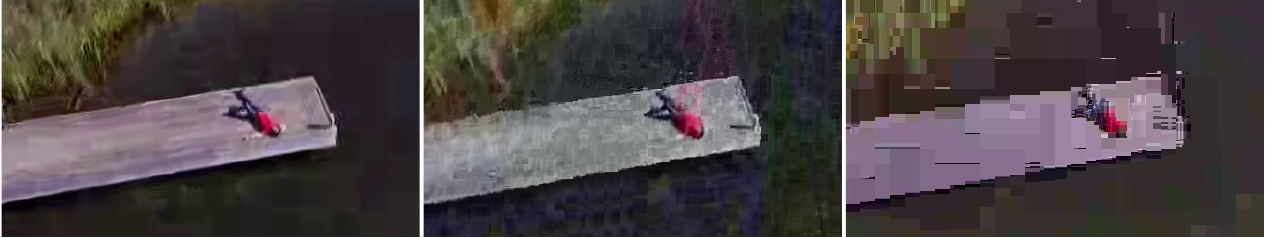}
   \protect\caption{Example frames from the evaluation sequence showing degradation of image quality when using different video encodings and a low bitrate of 50kbps, from the right: H.265, VP9, H.264. }
   \protect\label{fig:example_4_imgs}
\end{figure}

Even though existing mobile processing solutions can achieve acceptable performance levels, complementing the on-board computations with off-board processing can be very beneficial. In such cases, wireless transmission of video signals is inevitable in order to transport data to remote processing units. However, this comes at the price of increased delays and the introduction of compression artifacts, especially if the available bandwidth is limited. Typical examples of quality degradation in this setting, such as macroblocking, can be seen in Figure~\ref{fig:example_4_imgs}. 

\begin{figure}[t]
   \centering
   \includegraphics[width=1.0\columnwidth]{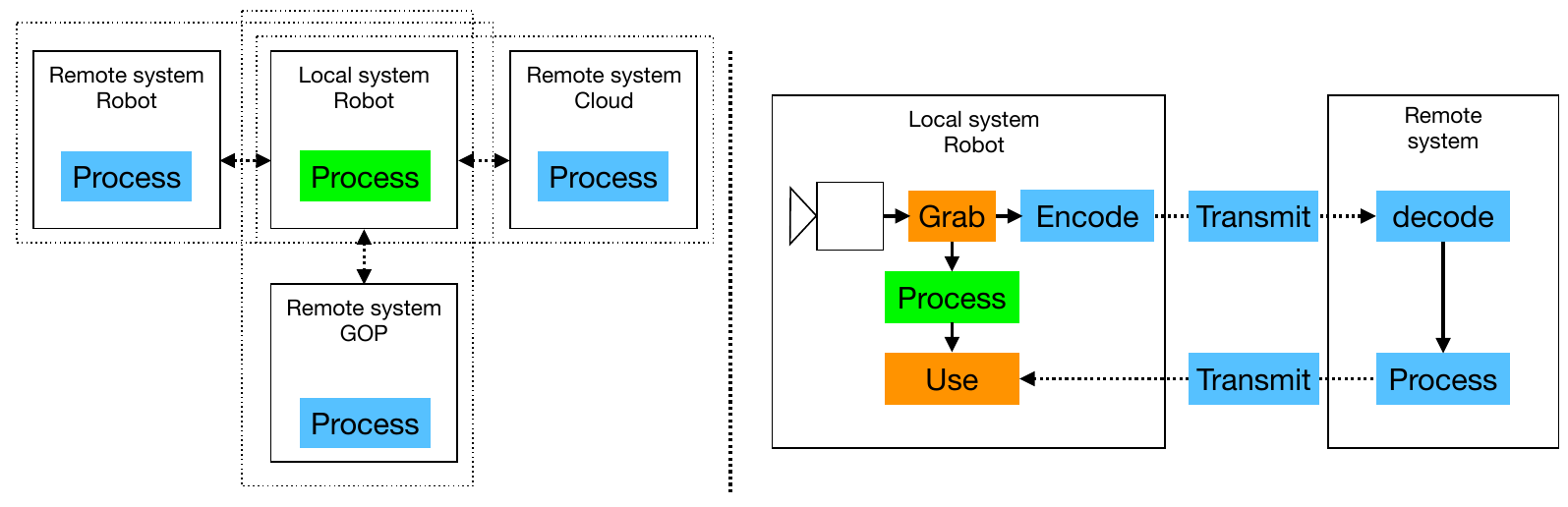}
   \protect\caption{Overview of the system components related to local versus remote processing. Left: computational resources available to a UAV platform. Right: processes involved in local (orange and green) and remote processing.}
   \protect\label{fig:local_vs_remote}
\end{figure}

The combination of autonomous UAVs, embedded computer vision algorithms, Cloud Robotics, and highly distributed teams of collaborative systems introduces additional complexities. In particular, the encoding, transmission, and decoding phases of the image processing tasks that may be distributed must be considered, as shown in Figure~\ref{fig:local_vs_remote}. It depicts the essential components involved in enabling remote processing used when operating in the field. The left part of the figure depicts several different computational sources available for a specific UAV to leverage. The right part of the figure provides the configuration that will be studied in this paper, where a local robotic system interfaces with remote systems with computational resources that, in this context, could be the Cloud, another UAV, or a GOP.

The proposed method described in this paper for taking advantage of different types of vision-based detectors does not limit the types of detectors used. Even though the focus is placed on the latest developments in the field in the form of deep convolutional networks, other types of algorithms can also be incorporated. The only prerequisite is quantifying the relative performance of a specific method within the pool of available techniques. One crucial issue to note is the execution time of a detector. CNN-based detectors are characterized by a constant (or at least very consistent) execution time independent of the input image. In the case of other techniques, this might be different. When it comes to quantifying the relative performance of algorithms, the focus so far has been placed on the degradation in performance under limited bandwidth video transmission. However, other criteria can be considered that can be more contextual. For example, one can select detectors based on the expected object size, orientation, or camera exposure. These additional criteria will be a subject of future work. 

From the perspective of the executed flight paths, we focus on the initial phases of a rescue operation. For this reason, the planned paths of the UAVs follow a scanning pattern that ensures complete coverage of an area of interest. This means that observations of a particular location can potentially be made only once and for a limited time. This initial detection of objects is performed without any feedback that could influence the flight paths. After this initial phase, a second flight can be performed to verify previously detected objects or improve the geolocation accuracy. Alternatively, other exploration strategies can be used in the initial phase, and the described method should still be applicable. However, this is a task for future work.

\subsection{Contributions}

 This paper investigates the use of deep convolutional neural networks for object detection and geolocation in the context of emergency rescue using aerial robots. In particular, the performance of several existing neural networks under various video compression techniques is evaluated. The degradation of object detection performance when dealing with video compression at low bitrates and the timing properties of the components involved while taking advantage of remote processing, including computation in the Cloud, is evaluated. TThis characterization allows for creating a pool of detectors and the ability to select them to process images optimally throughout a mission. The results of detections are probabilistically fused, which allows for the building of a saliency map where objects of interest become geolocated salient locations. The techniques have been integrated with the overall software architecture and used onboard UAV systems in real-world experimental missions.

The main contributions presented in this work are the following:
\begin{itemize}
   \setlength\itemsep{0em}
    \item Evaluation of several Deep Convolutional Neural Networks using Average Precision and Recall as well as Localization-Recall-Precision metrics using three video encoding techniques with limited encoding bandwidth. 
    \item A formal model for the object detector allocation problem that considers computational and communication resources.
    \item An Object Detector allocation algorithm for optimal selection of detectors that considers the available detectors and computational and communication resources. The optimization is based on the evaluation results provided as input to the algorithm in addition to the use of Integer Linear Programming as part of the algorithm to capture the complexity of pragmatic system constraints associated with in-the-field allocation of detectors.
    \item A novel method for probabilistic fusion of detection results, geolocation of objects, and generation of saliency maps of identified objects is specified.
    \item Integration of the techniques described above in the system architectures for UAV quadrotor systems, in addition to field test experiments made to validate the approach. 
\end{itemize}{}

\subsection{Structure of the paper}

The paper is structured around the main contributions. In Section~\ref{sec:related-work}, related work relevant to the topics studied is presented.
In Section~\ref{sec:networks}, deep network structures are reviewed, three families of networks are introduced, and the networks used for evaluation are described. Information is also provided about the datasets used, the evaluated compression techniques, and the results of the evaluations. In Section~\ref{sec:det_choice}, an algorithm is presented for optimally selecting object detectors that considers both the computational and communication resources. In Section~\ref{sec:fusion}, a method for fusing image-based detection results, geolocating the detected objects, and building a common map of salient object locations is presented. The results of experimental validation for the approach are provided in Section~\ref{sec:experimental}. Conclusions are provided in Section~\ref{sec:conclusions}.

\section{Related work}\label{sec:related-work}

This section focuses on a body of topics and representative related work relevant to the proposals in this paper. 

The use of UAVs for search and rescue applications has been steadily increasing in the past years in many settings. Early examples include searching in the wilderness~\cite{goo:07} or using larger scale UAVs (100kg) with color and thermal cameras in an urban setting~\cite{rud:08b}. Other domains include maritime SAR using UAVs and also include the use of surface vehicles~\cite{gal:18,yan:20}. \cite{lyu:23_survey}, provides a recent survey on the topic of search and rescue using drones.

A system description and its performance in a simulated but real-world experiment involving the use of UAVs for Search and Rescue has been described in~\cite{nie:18}. In this approach, a fixed-wing platform has been used to assist in a search for missing people in various terrain types. The data analysis component of the system takes advantage of a nested \emph{k}-means approach to automatically process images obtained by the platforms, including the computation of geographical coordinates of possible locations of a lost person~\cite{nie:17}. Although the method does not take advantage of deep convolutional neural networks, it has proven to be useful in real-world operations. 

The SHERPA project focused on the development of a heterogeneous robotic system combining ground and aerial robots collaborating with human operators to support alpine SAR missions~\cite{mar:13}. One of the results of this project was the design of a quadrotor platform UAV for Search and Rescue use in post-avalanche events~\cite{sil:17}. The system takes advantage of close-range ARTVA technology to detect victims buried under snow. This is an example of a non-vision based approach to assisting operators in rescue operations.
Considerable effort has been put into developing frameworks for dealing with information storage, synchronization, sharing, and use in the context of collaborative robotics applications. Concepts for the specification and prototyping of a general distributed system architecture that supports the creation of Hastily Formed Knowledge Networks (HFKNs) by teams of robots and humans have been presented in~\cite{doherty2021hastily}. The information collected by the team of agents ranges from low-level sensor data to high-level semantic knowledge, the latter represented in part as Resource Description Framework (RDF) Graphs~\cite{LASSILA-TR-1999}. The proposed framework includes a synchronization protocol and associated algorithms that allow for the automatic distribution and sharing of data and knowledge between agents. This is done through the distributed synchronization of RDF Graphs shared between agents presented in~\cite{berger2023rgs}. In~\cite{ir.2024.06}, an active query interface to complex multi-agent systems consisting of teams of heterogeneous robotic systems has been presented. Rather than manually and tediously setting up detailed missions, human operators can instead query a system, where the query is a higher-level declarative description of what is required to enhance situation awareness. An active query can automatically initiate many different agent processes to answer it, such as generating and delegating tasks to robotic team members (using delegation framework~\cite{doh:13}) required to achieve the specific goals implicit in the query.






From the perspective of object detection techniques used in SAR operations, two main trends exist. The first one takes advantage of \emph{traditional} methods, which do not use CNNs but might use other forms of machine learning, such as boosting. Even though the deep-neural network-based approaches outperform the traditional techniques in most applications, the fact that considerably less computational power is required for alternative methods is still advantageous, especially for small-scale platforms and on-board processing. Among these techniques, prevalent methods make use of Haar-like features~\cite{vio:01} and Histograms of Oriented Gradients (HOG)~\cite{dal:05}, either directly or with variations of these approaches.
Examples of the latter traditional approaches include ~\cite{blo:13,blo:14,agu:17a,agu:17b}. Examples of other non-CNN approaches (or combinations of techniques and/or hybrid approaches) include the use of human body pose estimation for suspicious behavior detection~\cite{pan:14} and making use of visual saliency to focus the search for objects on promising image regions~\cite{sok:10,zha:13,got:16}. Other methods take advantage of simpler techniques such as blob detection, image thresholding, or background subtraction ~\cite{gii:15,sun:16,vem:15,hua:10}. 

The most commonly used approach to object detection is using the visible light spectrum in the form of color images. However, many techniques also use infrared or thermal images ~\cite{mil:08,gas:11,fly:13,por:14,lei:15}.
Using this modality is advantageous, as it relies on the fact that objects such as human bodies or vehicles are often distinguishable from the background based on temperature difference. A survey of fusion methods and applications dealing with visible and infrared images can be found in~\cite{ma:19}.


The second trend is the use of CNN methods for UAV-based object detection has become more prevalent. Taking advantage of the rapidly developing body of work for this kind of application is being extensively investigated, and many CNN techniques are being adapted to the specific requirements of this application area. Based on these recent developments, detecting objects of different classes in images can be solved reliably and robustly on a wide range of hardware.

Several successful neural network structures have been developed and deployed in various contexts~\cite{ssd:16,red:16}. These networks are often trained and evaluated on several datasets, such as Common Objects in Context (COCO)~\cite{COCO:14} and are available for use in an off-the-shelf manner. Additionally, depending on the task at hand, a new detection neural network can be trained from scratch~\cite{liu:21}, or through transfer learning to suit a specific application area. For example, a number of network structure improvements are proposed in~\cite{ye:23} in order to deal with the problem of small object detection from UAVs. Similarly, UAV-YOLO~\cite{liu:20} proposes a number of improvements to YOLOv3 for the same purpose. In~\cite{gasienica2021ensemble}, the authors propose an ensemble model that combines multiple CNNs coordinated by the weighted majority voting fusion module for marine SAR operations. Similar to the approach presented in this paper, the proposed method aims to increase detection precision. However, this work considers additional practical runtime constraints, such as the available communication and computational resources. A comprehensive review of state-of-the-art deep learning-based object detection algorithms in low-altitude UAVs is presented in~\cite{mit:20}. 
A survey of object detection techniques, including their history, can be found in~\cite{aga:18_det_surv_1}. The field is very active and the number of publications doubled from 2011 to 2018 and continues to grow~\cite{zou:19_det_surv_2}.









Calculating the geographical coordinates of an object detected in an image can be achieved in various ways and has been the subject of many research endeavors.
For example, a method for improving geolocation accuracy using a ranging sensor has been described in~\cite{liu:17}. A method for relaxing the common assumption of a flat ground, which uses Digital Elevation Models (DEM), has been proposed in~\cite{qia:18}. In~\cite{zha:22}, the authors propose an approach that combines a continuous stream of images and GPS data collected using a UAV platform to generate a 3D geographic map and accurately estimate the GPS location of the target center pixel. These methods focus on detecting and tracking objects, but information about observed areas of the environments where no detections are made is ignored. This is in contrast to the method proposed in this paper, where this important kind of information is also captured when the saliency map is updated (see Section~\ref{sec:fusion}).

A method for fusing classifier observations using Gaussian Processes for the observation model has been proposed in~\cite{tea:15}. The approach does not focus on obtaining target locations but instead focuses on improving the probability of detecting static objects by correlating classifier outputs as a function of UAV position and searching for targets. 
Another approach for fusing collected information uses Sensor Fusion Quality (SFQ)~\cite{kwo:11}. The authors characterize the quality of a set of newly acquired sensor data containing a target by examining the joint target location probability density function.




The idea of connecting robots using wired or wireless networks in order to take advantage of their heterogeneity while solving tasks has been previously researched under the name \emph{Networked Robotics}~\cite{net_rob:00}. The name refers to a set of robots working in a coordinated and/or cooperating fashion and sharing data in order to solve a task. Limitations of individual robots, however, contribute to the collective limitation of the networked system and introduce new challenges. 
For example, even though the collected data, such as maps, can be shared between robots, the overall amount of information within the networked system is limited to the data collected by its individuals.
The same applies to the amount of available computational and storage resources. Even though they can be shared, they remain a constant quantity within the networked system. 

Due to advancements in the fields of big data and Cloud computing, the idea of \emph{Cloud Robotics} has emerged to address some of the limitations mentioned above. The term \textit{Cloud Robotics}, coined by James J. Kuffner in 2010~\cite{kuf:10}, refers to any number of robots or automation systems which take advantage of a Cloud infrastructure for managing data as well as execution of program code. In other words, Cloud Robotics is a system where all sensing, computation and storage is not only performed within standalone systems, but also by leveraging external computational and storage capabilities.

Inclusion of a Cloud infrastructure brings elastic and on-demand access to computational and storage resources which allows for dealing with, for example, enormous amounts of data produced by sensors. Moreover, the Cloud offers access to data and knowledge databases as well as models, which can be both machine- and human-made. The latter is done through crowdsourcing. Along with this on-demand access to data, models and code also become available. It is important to notice that, as soon as available, access to new data or improved algorithms can transparently be offered to a robotic system within a Cloud Robotics setting. This allows for increasing the capabilities of a Cloud Robotics system without making physical changes to individual robotic platforms participating in the system. 

A comprehensive survey of research within the field of Cloud Robotics can be found in~\cite{cloud_rob_surv:2018}. It provides an insight into system architectures and Cloud Robotics applications. Additionally, the authors identify a number of challenges, one of which is related to reducing communication delays when transferring data over, to, and within the Cloud and specifically focuses on data offloading, pre-processing, and compression techniques. The networking aspect is an unavoidable result of enabling remote access within networked as well as Cloud Robotics fields.
An example that takes advantage of Cloud computation for image processing and UAVs has been recently presented in~\cite{lee:17}. The authors describe a system which first performs a local check of \emph{objectness} to find potentials for detecting objects. More computationally intense algorithms for object detection are then performed on a remote computer in the Cloud. Substantial speed up of computation is reported due to the use of this additional powerful hardware. 

The area which deals with offloading computation has been studied from many perspectives. A survey of relevant work has been presented in~\cite{kum:13}. It provides background, introduces different techniques, and systems in the area of offloading computation.

Although the approach to allocating detectors presented in this work deals with the process of using the algorithms (inference), there exists work which deals with efficiently scheduling the training process ~\cite{che:23}, which is a potential future work direction for our approach.

An interesting trend in supporting rescue operations is crowdsensing. It involves large numbers of mobile devices processing and sharing information for a common purpose, for example, map building. While in the original formulation it used mobile phones or similar devices carried by people, the same principles can be applied to robots. An example of this trend involving UAVs is SWARM (Search With Aerial RC Multi-rotors)\footnote{\url{http://sardrones.org}, accessed: 2024}. It is an organization of thousands of volunteer UAV pilots offering assistance with search operations. 

An extension of this idea, which involves automation of processing tasks, has also been studied. For example, CrowdVision~\cite{lu:19_crowdvision} aims at taking advantage of local (on-device) processing and using offloading of computation when needed and appropriate. Similarly, the work of~\cite{whe:23} considers the problem of optimally assigning local and remote resources, leveraging a dual-path Network Utility Maximization framework and a hit-ratio estimator.

In summary, the work presented in this paper touches on several research fields that are usually studied in isolation. Our approach combines several techniques to solve the task at hand in an end-to-end fashion. It allows for taking advantage of progress in several fields but also imposes specific limitations on the techniques used. For these reasons, a direct comparison to other work is not straightforward.

\section{Evaluation of vision-based object detectors}\label{sec:networks}

The proposed approach toward developing an optimized selection mechanism for vision-based detectors begins with an application-specific evaluation of a collection of vision-based detection algorithms. The evaluation aims to provide a basis for developing an optimized selection mechanism for a pool of such algorithms that can be used for inference (i.e., performing detections) online during search and rescue missions. The available algorithms are assessed using a set of metrics to determine their relative rankings. Additionally, the execution timing performance of these algorithms on the targeted hardware is evaluated. These results were partially presented in~\cite{rud:19}.

In this section, a procedure for algorithm evaluation is introduced, various network architectures are described, and a dataset representative of the targeted application used for evaluation purposes is described. The evaluation results regarding the accuracy and timing of different hardware configurations are then presented. 

\subsection{Detector evaluation procedure}

Algorithm~\ref{alg:eval_protocol} outlines the procedure for creating a set of execution characteristics for an object detector. The initial step involves selecting a dataset that represents the application at hand. As explained in Section~\ref{sec:intro}, each application area has unique properties that influence object sizes, occlusion, and other factors. Therefore, the evaluation dataset should encompass examples that accurately reflect the specific characteristics of the target application. Additionally, the dataset should include all the relevant object classes for the application. The annotation process for ground truth detections can be time-consuming, but it is only done once. The evaluation results are then used as a basis for an object detection selection mechanism.

\begin{algorithm}[t]
\KwData{detector to evaluate, evaluation dataset}
\KwResult{characterization of the detector}
  initialization\\ \label{eval:line:init}
\ForEach{inference setup}{\label{eval:line:inf_setup}
  \ForEach{image in the dataset}{\label{eval:line:image}
    execute the detector on the image\\ 
    measure the execution time
  }
  compute execution statistics\label{eval:line:compute_stats}\\
  \ForEach{metric}{\label{eval:line:eval}
    evaluate the obtained detections using the metric
  }
  collect the execution characteristics and performance metrics
}
collect complete detector characteristics 
\caption{Detector evaluation protocol}\label{alg:eval_protocol}
\end{algorithm}

After the initialization step meant for housekeeping (line~\ref{eval:line:init}), the evaluation procedure is performed for each inference setup (line~\ref{eval:line:inf_setup}). It is comprised of a combination of the host hardware and the software used for performing the inference. The hardware can be a physical standalone system or a cloud-based system, or any other inference service which can execute a specific detector and provide detection results. The execution conditions used during the evaluation should reflect the on-line conditions when the inference will be performed in order to be able to rely on the timing evaluation characteristics during actual missions. In the case that this cannot be reliably done, a margin of error can be applied to reflect the worst-case execution timing. The inference software setup can use any number of frameworks, such as  TensorFlow~\cite{tensorflow:15}, PyTorch~\cite{pytorch:17}, or other methods. 

For every inference setup, the procedure starts by executing the chosen detector on each image in the evaluation dataset (line~\ref{eval:line:image}). The results, in the form of detection bounding boxes and confidences, are collected along with the execution time for each frame. The latter are used to compute overall characteristics in terms of minimum, maximum, and average execution times (line~\ref{eval:line:compute_stats}). The following step computes all the relevant evaluation metrics (line~\ref{eval:line:eval}). These can include any number of methods for evaluating detection performance. The calculation is performed using the detections obtained in the previous step as well as the ground truth bounding boxes of the evaluation dataset.

The result of the evaluation procedure of a specific detector is a set of properties which describe its performance relative to a number of metrics and execution properties using a number of inference configurations. A complete set of evaluations performed on all available object detector algorithms results in a pool of object detectors from which all or some can be instantiated at runtime during a mission based on the specific mission context. For example, using a method described in Section~\ref{sec:det_choice}.


\subsection{Network architectures and configurations}\label{sec:net_architectures}

The purpose of detecting objects in images (using deep neural networks or otherwise) is to find bounding boxes of objects belonging to specific classes, including detection confidence measures for each box. In a manner similar to~\cite{hua:17}, our focus is on three families of detection networks which encompass several specialized variations. Single Shot Multibox Detectors (SSD)~\cite{ssd:16}, Faster Region-based Convolutional Networks (\mbox{R-CNN})~\cite{ren:17_faster_rcnn}, and Region-based Fully Convolutional Networks (\mbox{R-FCN})~\cite{dai:16} are evaluated. For different network meta-architectures, one of the following feature extractors is used: ResNet (50 and 101)~\cite{he:15_resnet}, Inception~v2~\cite{inceptionv2}, and Inception ResNet v2~\cite{sze:16:inc_res}. Short descriptions of the network architectures, as well as the feature extractors mentioned, are given below. In-depth descriptions can be found in~\cite{hua:17} and in the respective publications referenced above. 

\subsubsection*{Networks}
The following three meta-architectures have been evaluated:
\begin{itemize}
\item \emph{SSD}: Single Shot Multibox Detector architectures use a single feed-forward convolution network to directly predict classes and bounding boxes. By using a single stage, the need for a second stage where a number of proposals are classified, is omitted. This is in contrast to networks which work in two stages as described below. This approach discretizes the output space into a set of default boxes with different ratios and scales per feature map. At prediction time, the network computes scores for objects detected in each default box and adjusts the box so that its shape better matches the object in the image being processed~\cite{ssd:16}. The main advantage of SSD networks is that they can perform inference relatively fast, but on the other hand, do not work well with small objects.

\item \emph{Faster R-CNN}: Region-based Convolutional Networks (R-CNN) operate in two stages. First, a Region Proposal Network (RPN) is used to generate detection region proposals. In the second stage, the generated proposals (typically 300) are evaluated using a classifier. Faster R-CNN~\cite{sha:15} uses a network architecture which overcomes shortcomings of its previous versions: R-CNN and Fast \mbox{R-CNN}. Due to the speed improvements, Faster \mbox{R-CNN}s are suitable for real-time operation. More information can be found in~\cite{ren:15_faster_rcnn}.

\item \emph{R-FCN}:
Region-based Fully Convolutional Network~\cite{dai:16} structure is similar to Faster R-CNN, but the cropping of features occurs in the last layer prior to class prediction. This is unlike Faster R-CNN, where the cropping is done from the same layer where region proposals are predicted. This allows for a significant gain in speed as the amount of per-region computation is minimized.
\end{itemize}

\subsubsection*{Feature extractors}
In all the network families mentioned above, a convolutional feature extractor transforms an input image to a set of high-level features first. The following feature extractors have been chosen for the evaluation.
\begin{itemize}
\item \emph{ResNet (50 and 101)}: Training of very deep networks becomes difficult and the vanishing gradient problem is one of the reasons the accuracy starts to saturate and can even degrade. To overcome this issue, the layers have been reformulated as learning residual functions with reference to the inputs instead of learning unreferenced functions. Here, the residual can be understood as a subtraction of a feature learned from the input of the layer. ResNet structure also uses shortcut connections between layers. Due to these changes, the learning of networks of this form is easier than simple deep networks, and the issue with degrading accuracy is also resolved. The introduction of Residual Networks is considered one of the bigger breakthroughs in the area. The designation of 50 and 101 (also 152) refers to the number of layers in the network~\cite{he:15_resnet}.

\item \emph{Inception v2}: The choice of kernel size depends on the locality of the salient information (object of a class) in an image. For local information, a small kernel is better, and for global, a larger one is more suitable. Very deep networks are prone to overfitting. To solve this issue, it has been proposed to go \emph{wider} instead of \emph{deeper} - filters with multiple sizes are placed on the same level (1x1, 3x3, 5x5). This is known as GoogLeNet (Inception v1). Inception v2 increased speed and accuracy by factorizing 5x5 into two 3x3 and nxn to a combination of 1xn and nx1 to improve speed~\cite{inceptionv2}.

\item \emph{Inception ResNet v2}: It is a hybrid of ResNet and Inception networks which introduces residual connections that add the output of the convolution operation of the inception module to the input~\cite{sze:16:inc_res}. 
\end{itemize}

\subsubsection*{Configurations}\label{sec:net_configs}
\begin{figure}[t]
   \centering
   \includegraphics[width=0.95\columnwidth]{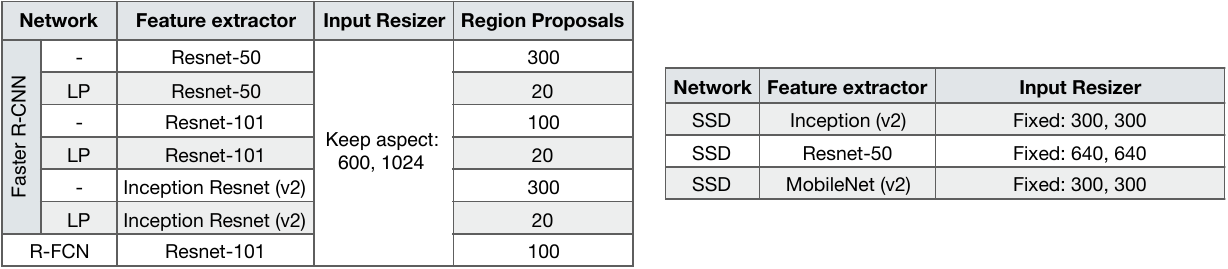}
   \protect\caption{Network configurations used for evaluation.}
   \protect\label{fig:net_configs}
\end{figure}

The table in Figure~\ref{fig:net_configs} presents the network configurations evaluated in this paper. In addition to the network family, the type of feature extractor is also specified. For all networks, the input image is first resized in one of two ways. Either it is scaled to a fixed size ($300\times300$, or $600\times600$ pixels) or it is resized such that the shorter edge is 600 pixels and the longer is scaled to be no longer than 1024 pixels. The scaling is done in a manner that preserves the aspect ratio. 
Finally, the number of produced region proposals is specified where applicable. A typical number of proposals evaluated by Faster R-CNN and R-FCN is 300. In order to speed up the detection process the number of proposals can be reduced with a small penalty on the recall performance, as reported in~\cite{hua:17}. Since our interest is in close to real-time performance,  the effect of using a lower number of proposals on the evaluation dataset has been investigated. For the evaluation, already trained networks available within the TensorFlow~\cite{tensorflow:15} models zoo have been used~\footnote{TensorFlow 1 detection model zoo: \url{https://github.com/tensorflow/models/blob/master/research/object_detection/g3doc/tf1_detection_zoo.md} (2024)}. 

\subsection{Evaluation dataset}\label{sec:dataset}

In order to evaluate the degradation of performance of object detectors due to image compression, a number of aerial sequences have been selected from videos collected during a two-year period. The selection was made to assure variations in weather conditions, times of year, locations, detection subject's poses and appearance (clothing), and backgrounds (grass, gravel, wood, asphalt, dirt road, water). A total of 69 minutes of source video footage has been used. 
 
\begin{figure}[t]
   \centering
   \includegraphics[width=0.9\columnwidth]{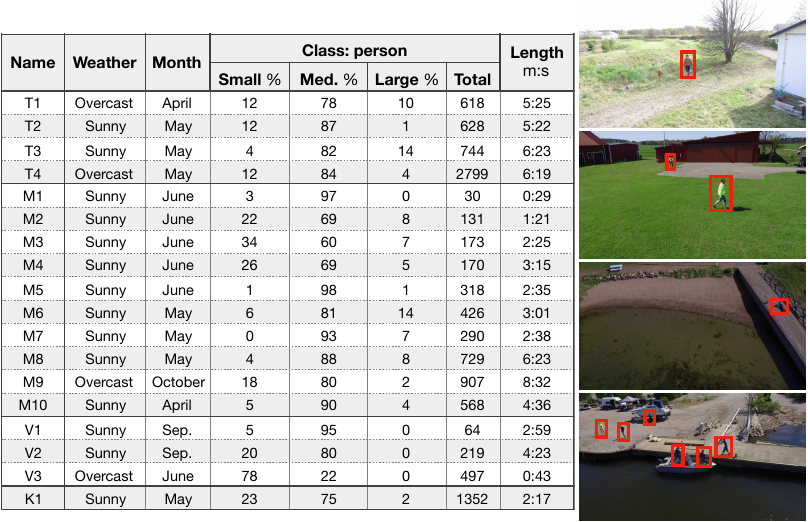}
   \protect\caption{Summary of video sequences and example images used for evaluation: four locations (T, M, V, K), different weather conditions and times of year, percentage size composition (small, medium, and large) and total number of objects of class person.}
   \protect\label{fig:vid_sequences}
\end{figure}

The table in Figure~\ref{fig:vid_sequences} presents descriptions of the sequences collected at four locations abbreviated as T, M, V, and K. The right side of the figure shows a number of sample images from these locations. The distance of the camera and objects to be detected is characterized by the percentage of objects belonging to groups defined in the COCO evaluation protocol: small (area\textless$32^2$pix), medium ($32^2$pix\textless area\textless$96^2$pix), and large (area\textgreater$96^2$pix). 

The original sequences were recorded using  DJI Zenmuse Z3 cameras on  DJI Matrice 100 and  Matrice 600 Pro platforms. The videos were recorded on the cameras' internal SD cards with $1920\times1080$ resolution and a 60Hz frame rate using the H.264 codec with a bitrate of 60Mbps. In the preparation of the evaluation sequences, the originals were transcoded using the codecs described below in Section~\ref{sec:compression} to a resolution of $1024\times576$ and a 30Hz frame rate. This resolution and frame rate were chosen as they are more common for cameras typically used on UAV platforms for on-board processing, as well as to fit to the inputs of the neural networks evaluated. A wide range of bitrates have been used, spanning from 20Mbps, down to the extremely low 50kbps. 

For SSD networks which use fixed-size input images with an aspect ratio of 1:1, no additional pre-processing has been applied. It has the disadvantage that the input images are deformed when scaled from the original 16:9 aspect ratio. To remedy this, it is common to crop the images or add padding in order to preserve the aspect ratio. 

Finally, images at the rate of 2Hz were extracted from the transcoded sequences to produce more than 8200 images per codec and per bitrate. Because we are dealing with video sequences, unlike the case of typical image data sets, the consecutive images are similar. For this reason, images at a smaller rate are still representative of the full sequences. Finally, the evaluation images were annotated with the \emph{person} class and used as ground truth during evaluation. 

\subsection{Video compression techniques}\label{sec:compression}

%

Three common video compression techniques were investigated in this work: H.264, H.265, and VP9. The H.264 (MPEG-4 AVS) coding format is currently one of the most commonly used techniques for a multitude of applications. Compared to its predecessor (MPEG-2 Part 2), it offers bitrate savings of up to 50\% with comparable quality. The H.265 (High Efficiency Video Coding - HEVC) is a successor to H.264 and is quickly gaining market share. Again, it offers up to 50\% of the required bitrate reduction as compared to H.264. The third codec under evaluation is an open, royalty-free media file format called VP9. It offers bitrate savings of up to 20–50\% when compared to H.264. For more details about the coding techniques, see for example~\cite{guo:18}.

The encoding configurations for the application considered in this paper have been chosen to minimize the encoding time, that is to allow to transfer, decode and process the data with minimum delay. The evaluation and results characterize the degradation and performance of object detection, where minimal time increase is paramount. For applications where timely delivery of results is not a priority (for example off-line and/or batch processing), different encoding configurations can be more suitable. This would result in either better image quality at the same bitrate or decreased required bitrate for the same quality. The results presented below correspond to the \emph{worst case}, which is the lowest quality settings with minimal delays. 

For the purpose of evaluation, the codecs were used in the following way. Video sequences described in the previous section were encoded using the FFMpeg\footnote{FFmpeg multimedia framework: \url{https://www.ffmpeg.org}} multimedia framework. The most important parameters were chosen as follows: \emph{1-pass} encoding with a target bitrate and \emph{ultrafast} preset. The latter parameter allows for achieving a very fast encoding time, but the resulting compression level is not optimal. 

It is worth mentioning that a relatively recent codec called AV1\footnote{Alliance for Open Media: \url{https://aomedia.org}} is intended to supersede the current ones in the coming years. Its adoption should be faster as AV1 is a royalty-free codec. It promises a 30\% bitrate reduction as compared to H.265 or VP9. However, at the time of writing, it is in its early stages of adoption. The available implementations are experimental and the support as well as performance of encoding and decoding are limited and available only on the newest hardware (GPUs and CPUs). 

\subsection{Detection performance evaluation}\label{sec:perf_eval}

In this section, first, suitable metrics for evaluating object detectors' performance are discussed and identified.
This identification lays the basis for making an optimal choice of object detectors that takes into account the practical limitations that apply when operating in the field, given the requirements associated with the particular mission context. 
Secondly, the performance evaluation results are presented using these metrics under limited bandwidth with different encoding techniques. 

With the search and rescue target application in mind, there are two important aspects to evaluate: (1) the ability to minimize the number of false negative detections during a search mission, and (2) being able to accurately geolocate positions of correctly detected objects. The first aspect deals with the \emph{classification} accuracy, which involves analyzing recall and precision. The second aspect deals with the accuracy of \emph{localization} of objects in the image. This translates into the geolocation accuracy, which is performed at a later stage of the overall method. The two properties might have different priorities depending on the mission at hand and it is important to be able to make the decision of which detectors to use based on these factors.

Many different metrics have been proposed for evaluating object detectors. The most prominent in the literature is the Average Precision (AP), which in short is the area under the recall-precision curve. Variations of this metric are the most commonly reported in the literature and are used for evaluating object detection algorithms. This common and standard metric allows for comparing incremental improvements of algorithms reported in the literature. Another metric has been proposed, which has interesting properties from the perspective of the task focused on in this paper. Localization-Recall-Precision (LRP)~\cite{oks:18} is a relatively new metric specifically designed for object detection. The LRP error is composed of three components related to: localization, false negative (FN) rates, and false positive (FP) rates. Below, we briefly introduce the two metrics.




Table~\ref{tab:bin_classification} shows the binary classification contingency matrix adapted to object detection.

\newcommand{\binsize}{\resizebox{0.8\width}{!} }
\begin{table}[htb]
  \begin{center}
\binsize{
  \begin{tabular}{ | c | c | c | } 
  \hline
  \chl \textbf{}&\chl \textbf{Object Present}&\chl \textbf{Object Absent}\\
  \hline
 \chl  \textbf{Detection Positive}&True Positive (TP)&False Positive (FP)\\
  \hline
 \chl  \textbf{Detection Negative}&False Negative (FN)&True Negative (TN)\\
  \hline
  \end{tabular}
  }
  \caption{Binary classification table.} \label{tab:bin_classification}
  \end{center}
\end{table}

The table defines the combination of cases (TP, FP, FN, TN), for when an object exists in an image and the result of object detection. Using these terms, Precision and Recall are defined by the following equations:
\begin{equation}
Precision=\frac{TP}{TP + FP}\\
\protect\label{eq:precision}
\end{equation}
\begin{equation}
Recall=\frac{TP}{TP + FN}\protect\label{eq:recall}
\end{equation}

In machine learning, \emph{Precision} answers the question: what proportion of positive detections was actually correct? (top row in Table~\ref{tab:bin_classification}). In other words, a model that produces no false positives has a precision of~1 which means that all positive detections were made for existing objects. Similarly, \emph{Recall} answers the question: what portion of actual positives (an object is truly present) was identified correctly? (left column in Table~\ref{tab:bin_classification}). A model that produces no false negatives has a recall of~1 which means that all the present objects were detected. The two parameters together describe the behavior of a detection algorithm. In many applications, there exists an inverse relationship between precision and recall, so that it is possible to increase one at the cost of the other. 

\begin{figure}[ht]
   \centering
   \includegraphics[width=0.1\columnwidth]{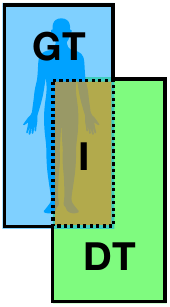}
   \protect\caption{Intersection over Union definition as a relation of the Ground Truth (GT) and a Detection (DT): the area of the intersection I (dashed line) divided by area of the union (solid line).}
   \protect\label{fig:iou}
\end{figure}

For object detection where a bounding box is the output, the decision whether a detection should be considered a True Positive (TP) is determined by checking the relation between Ground Truth (GT) and Detection (DT) (prediction) bounding boxes. This is schematically presented in Figure~\ref{fig:iou}. The Intersection over Union (IoU) is defined as the area of the overlap (intersection) divided by the area of the union of the two bounding boxes. If the bounding boxes of DT and GT are equal, the value of IoU becomes 1. If no overlap exists between DT and GT, the value of IoU is 0.
Commonly, a threshold on IoU is defined above which a prediction is considered a TP (given that the classification is correct as well). If IoU is below the threshold, the classification is considered a FP (also for duplicate bounding boxes). A detection is considered a FN if no detection is predicted for a given ground truth box or the predicted class was incorrect. 

Average Precision (AP) is defined as the area under the plot of the recall-precision (RP) curve. It is created by going over the list of detections sorted by the confidence score and finding the value of precision for each recall value. The values of precision and recall are calculated using Equations~\ref{eq:precision}-~\ref{eq:recall} at a specific IoU threshold (e.g. 0.5 denoted AR@0.5). Different variations exist in various AP-based metrics for how the "area under the curve" is calculated. For example, in PASCAL VOC2010–2012~\cite{eve:09_pascal_voc}, the recall curve is sampled at unique recall values (when recall changes); in the older PASCAL VOC metric, the recall is sampled at fixed 11 points; and in COCO AP~\cite{COCO:14} the recall is sampled at fixed 101 points - from 0 to 1 at 0.01 steps. 

Additional designation of the AP metric which deals with the chosen IoU threshold has the following common values. It can be a single value of 0.5 (e.g. PASCAL VOC) or 0.75. For the COCO AP, there also exists an average of over 10 IoUs from 0.5 to 0.95 with a step size of 0.05 (denoted as AP@[.5:.95]). This average is calculated with the purpose of providing a single value which rewards detectors that are better at localizing objects in the image. Moreover, mean Average Precision (mAP) is an average over all evaluated classes (categories). However, mAP and AP are sometimes used interchangeably, depending on the context.

Average Recall (AR) summarizes the distribution of recall for IoU thresholds from 0.5 to 1 (AR@[.5:.95]). In COCO, only this range of IoUs is provided since detection performance correlates with recall at thresholds above 0.5. AR is defined as double the area under the recall-IoU curve: $AR=2\int_{0.5}^1 recall (IoU)dIoU$. While both AP and AR are closely related, it is the PASCAL VOC/COCO mAP that is most commonly used for evaluations and reported in the literature for object detection~\cite{zou:19_det_surv_2}.

For the target application in this paper, one wants to optimize for high recall (fewer false negatives) and accurate object localization, and therefore, the Average Recall is a more important evaluation metric. Additionally, the COCO Detection Evaluation protocol defines three object sizes: small (area\textless $32^2$), medium ($32^2$ \textless area \textless $96^2$), and large (area \textgreater $96^2$). Both AR and AP are provided for these cases. 

The second performance metric which is used in the evaluation is the Localization Recall Precision (LRP)~\cite{oks:18}. It is a recently introduced metric specifically designed for object detection. The LRP error consists of three components related to: localization (IoU), false negative rate (FN) and false positive rate (FP)\footnote{FN and FP are defined differently here to denote the metric's components, and $N_{FP}$ as well as $N_{FN}$ denote the number of false positives and negatives, respectively. We keep the original notations for clarity.}. For a given set of ground truth boxes $X$, predictions $Y$, and parameters: a score threshold $s$ ($0\leq s \leq1$) and a IoU threshold $\tau$ ($0\leq s <1$), the LRP error is calculated as follows. First, prediction bounding boxes $Y_s$ with confidence levels above $s$ are assigned to the ground truth boxes in the same manner as in the case of AP. The LRP error is defined as:
\begin{equation}
LRP(X, Y_s)=\frac{1}{Z}(w_{IoU}*LRP_{IoU}(X, Y_s)+ w_{FP}*LRP_{FP}(X, Y_s) + w_{FN}*LRP_{FN}(X, Y_s)),
\protect\label{eq:lrp} 
\end{equation}
with $Z=N_{TP}+N_{FP}+N_{FN}$, where the three $N_{*}$ denote the number of true positive, false positive, and false negative detections, respectively. The three weights: $w_{IoU}=\frac{N_{TP}}{1-\tau}$, $w_{FP} = \vert Y_s\vert$, and $w_{FN}=\vert X\vert$ control contributions of the respective terms. 

The three components of LRP are defined as follows: 

\begin{equation}
LRP_{IoU}(X, Y_s)=\frac{1}{N_{TP}}\sum_{I=1}^{N_{TP}}(1-IoU(x_i, y_{x_i})),
\protect\label{eq:lrp_iou}
\end{equation}
\begin{equation}
LRP_{FP}(X, Y_s)=1-Precission=1-\frac{N_{TP}}{|Y_s|}=\frac{N_{FP}}{|Y_s|},
\protect\label{eq:lrp_fp}
\end{equation}
\begin{equation}
LRP_{FN}(X, Y_s)=1-Recall=1-\frac{N_{TP}}{|X|}=\frac{N_{FN}}{|X|},
\protect\label{eq:lrp_fn}
\end{equation}

where $IoU(x_i, y_{x_i})$ is computed for the ground truth bounding box $x_i \in X$ and its assigned prediction bounding box $y_{x_i} \in Y_s$. The component in Equation~\ref{eq:lrp_iou} represents the IoU tightness of valid detections. Finally, the LRP error can be also expressed as follows:
 
\begin{equation}
LRP(X, Y_s)=\frac{1}{N_{TP}+N_{FP}+N_{FN}}*\Bigg(\sum_{i=1}^{N_{TP}}\frac{1-IoU(x_i, y_{x_i})}{1-\tau}+N_{FP}+N_{FN}\Bigg).
\protect\label{eq:lrp2}
\end{equation}

The three contributions are averaged by the sum of the $N_{*}$ and the TP component is penalized by the error of localization ($1-IoU$) normalized by $1-\tau$.

Optimal LRP is defined in the following way:
\begin{equation}
oLRP=\min_s LRP(X, Y_s).
\protect\label{eq:olrp}
\end{equation}
It is defined as the minimum achievable LRP error with $\tau=0.5$. The quantity finds the optimal setting for score threshold $s$ which balances precision-recall-IoU for a specific class. The components of oLRP are: optimal box location ($oLRP_{IoU}$), optimal FP ($oLRP_{FP}$), and optimal FN ($oLRP_{FN}$). In comparison to AP, the oLRP finds the best class specific setting $s$ for a detector that is the minimum achievable LRP error representing the best achievable configuration of the detector in terms of recall-precision and the tightness of the boxes. Additionally, according to the authors, the shortcomings of AP which oLRP improves upon are: (a) AP is not confidence score sensitive, (b) 
AP does not provide the best detector setting in the form of a confidence score threshold, and (c) AP uses interpolation between neighboring recall values which is problematic for classes with small sizes~\cite{oks:18}. 




\subsubsection*{AP and AR Evaluation}\label{sec:eval_AP}

The results presented below were obtained by performing evaluations using the COCO evaluation protocol. In order to provide the most insight into the performance degradation under limited bitrate for different codecs, results are presented in terms of most of the COCO metrics and comments made on the metrics that are omitted. The results were obtained by performing detections using the dataset described in Section~\ref{sec:dataset} and video encoding techniques described in Section~\ref{sec:compression}. 


\begin{figure}[ht]
   \centering
   \includegraphics[width=0.99\textwidth]{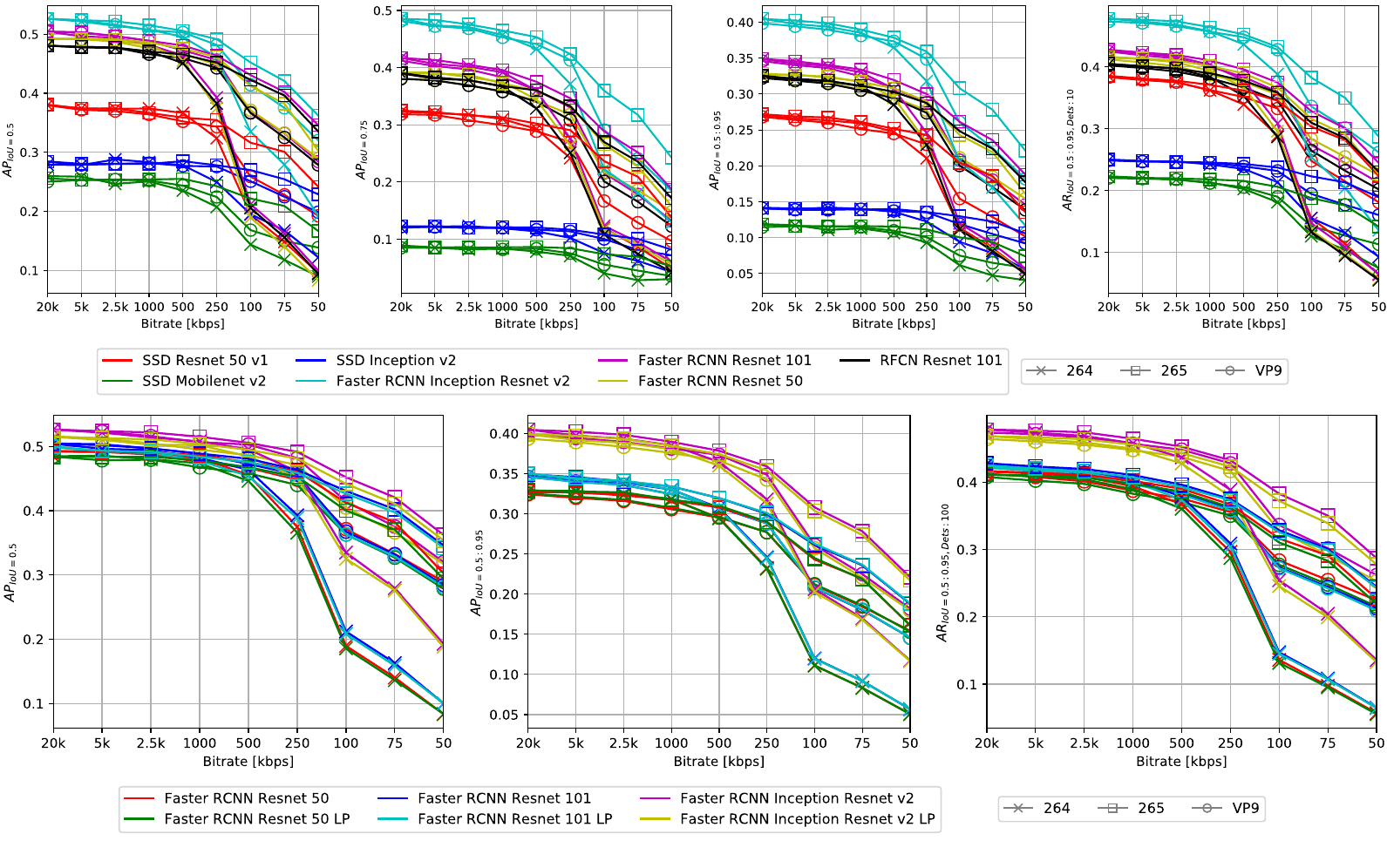}
   \protect\caption{Top: Average Precision for three IoU thresholds and Average Recall for different networks, codecs and bitrates. Bottom: Average Precision for two values of IoU and Average Recall of Faster R-CNN networks with lower number of proposals (LP).}
   \protect\label{fig:ap_ar_all_lp}   
\end{figure}

The top of Figure~\ref{fig:ap_ar_all_lp} presents Average Precision and Average Recall values for different networks, codecs, and bitrates. It shows the degradation of the detection performance as the video compression bitrate becomes more restrictive. The first three plots present the Average Precision for three IoU settings: 0.5, 0.75, and 0.5:0.95. The rightmost plot presents Average Recall at the IoU setting of 0.5:0.95 and a maximum of 10 detections. The setting with a maximum of 100 detections is omitted, as it is mostly identical to the 10 detections case for the dataset used. The setting of 1 maximum detection has a very similar shape as the case of 10. In general, all networks do not suffer from reducing the bitrate down to 2.5Mbps  or even 1Mbps for all encoding techniques. Below this value, the performance deteriorates rapidly for H.264 but more gracefully for VP9 and H.265, respectively. The relatively best performance is achieved using H.265 even when the bitrate is reduced down to 500kbps. The performance of the SSD Inception v2 and Mobilenet v2 networks is constantly lower relative to other networks but stays mostly constant for almost all bitrates. This behavior is expected as these two networks use input of smaller size, as presented in Figure~\ref{fig:net_configs}. If these networks are (for example, due to lack of appropriate hardware) the only available option, there is no need to devote additional bitrate for video transmission.

The bottom of Figure~\ref{fig:ap_ar_all_lp} presents the difference in average Precision and Recall for Faster \mbox{R-CNN} networks depending on the number of proposals used. The configuration of networks is presented in Figure~\ref{fig:net_configs}. The detection performance both in terms of AP and AR is very close for both cases when all proposals and the lower number is used. In the case of the dataset used for evaluation, the use of low-proposal networks is justified as it does not seem to deteriorate the performance and it offers big gains in computation times as described in Section~\ref{sec:eval_timing}. However, if the best object detection performance is required, a higher number of proposals is better. The general performance of codecs at specific bitrates and codecs follows the findings described previously.

\begin{figure}[ht]
   \centering
   \includegraphics[width=0.99\textwidth]{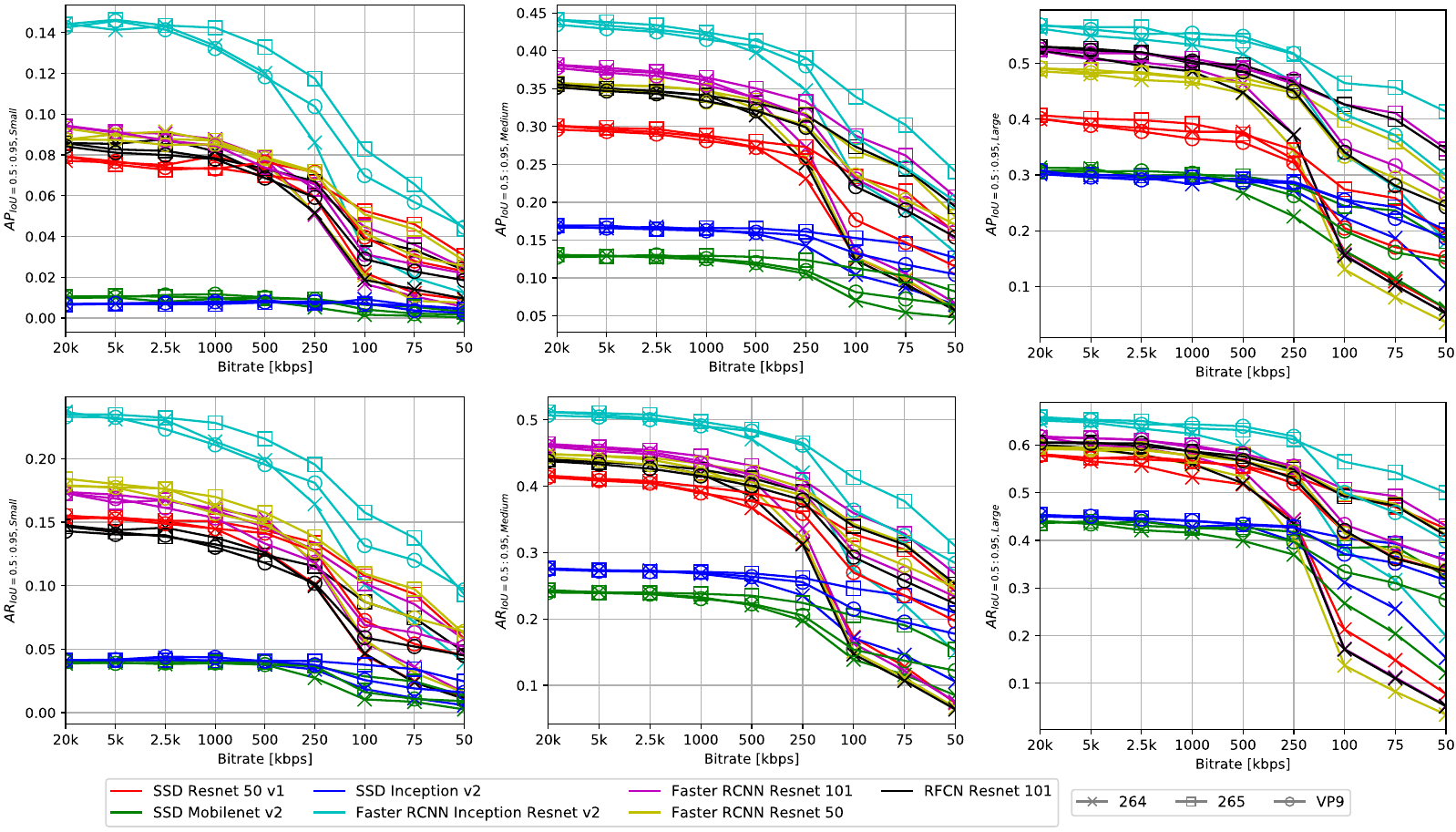}
   \protect\caption{Average Precision (top) and Recall (bottom) at IoU 0.5:0.95 for objects of different sizes: small (left), medium (middle), and large (right) for networks, codecs and bitrates.}
   \protect\label{fig:ap_ar_sizes}
\end{figure}

Figure~\ref{fig:ap_ar_sizes} presents the Average Precision (top row) and Recall (bottom) for objects of different sizes as defined in the COCO evaluation protocol for different networks, codecs, and bitrates for the IoU setting of 0.5:0.95. The percentage of images of different sizes in the evaluated dataset presented in Figure~\ref{fig:vid_sequences} has the following breakdown: small: 16\%, medium: 79\%, large: 5\%. The performance of detecting small images in terms of AP and AR (left of Figure~\ref{fig:ap_ar_sizes}) is much lower than medium and large examples. The best performance is achieved by the Faster RCNN Inception Resnet v2 network. The performance starts to deteriorate for bitrates lower than 1 - 2.5Mbps. The constantly poor performance is characteristic of SSD Mobilenet v2 and Inception v2 networks at all bitrates. The performance for detecting large objects (right of Figure~\ref{fig:ap_ar_sizes}) is the best, as expected. Even the performance of the two mentioned SSD networks is relatively good in this case. Since medium-sized objects constitute 79\% of the dataset, the main contribution of the characteristics of overall AP and AR comes from this type of detections.   

\subsubsection*{LRP Evaluation}\label{sec:eval_LRP}

\begin{figure}[t]
   \centering
   \includegraphics[width=1\textwidth]{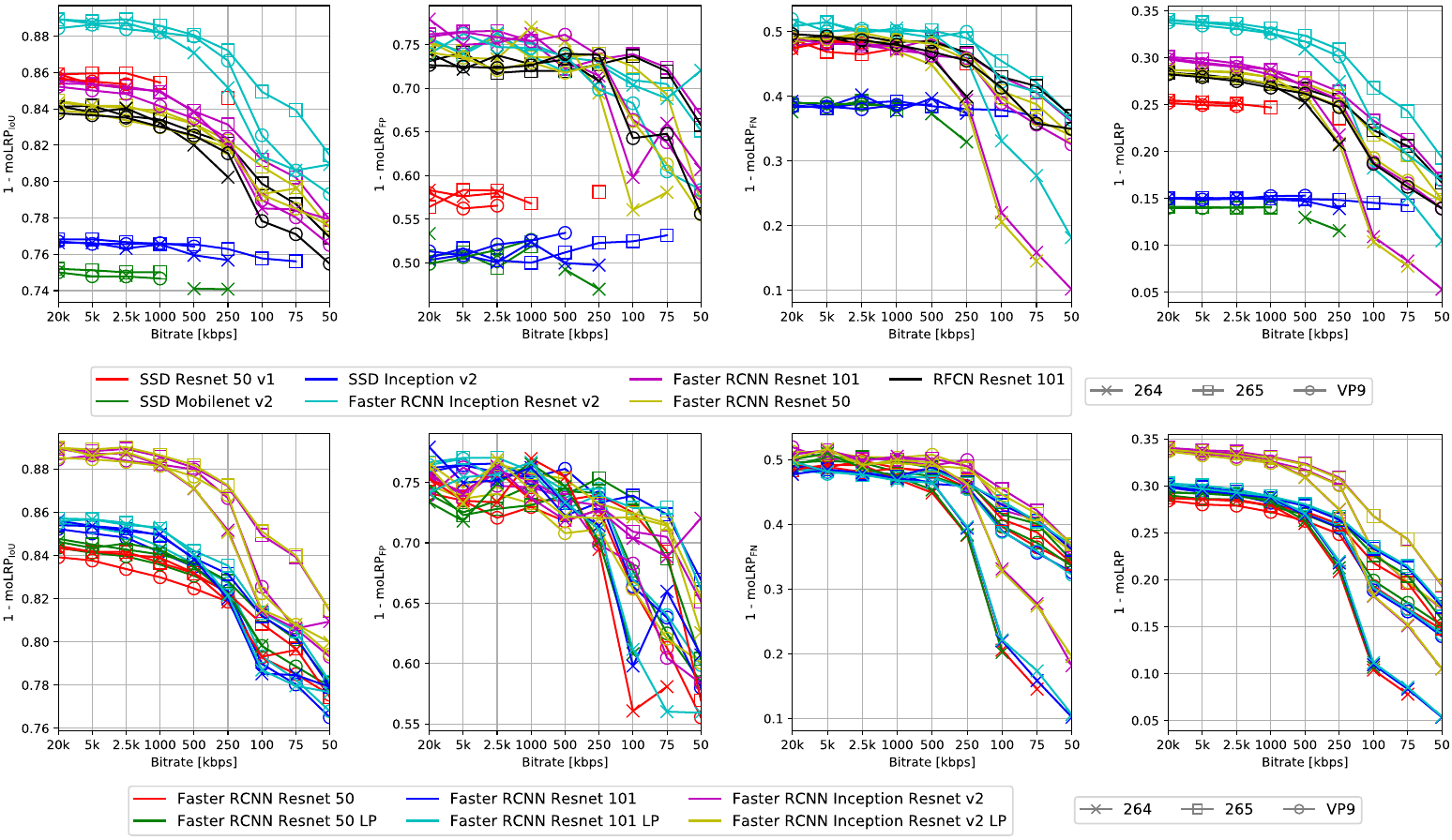}
   \protect\caption{1 - moLRP values for different networks, codecs and bitrates including components related to localization (IoU), false positives (FP), and false negatives (FN). Top row: networks without low-proposals versions. Bottom row: comparison of Faster R-CNN networks with lower number of proposals (LP). }
   \protect\label{fig:moLRP}
\end{figure}

Results of evaluation of the influence of bitrate limitation on the networks and codecs using the moLRP metric and its components\footnote{1: Note that we plot 1-moLRP values to make the plots visually comparable to AP and AR plots. 2: Since we show results only for one class: \emph{person}, moLRP and oLRP are the same as moLRP$=\frac{1}{\vert C\vert} \sum_{c \in C}^{}oLRP_c$, where C is a set of all classes, here C=\{person\}.} are presented in Figure~\ref{fig:moLRP}. The top row presents results for networks without low-proposal versions, while the bottom row shows Faster R-CNN networks with a lower number of proposals denoted with LP. For both cases, three components: IoU, FP, and FN, of the metric are presented as well as the combined value. 

In practical applications of vision-based detectors, it is often the case that a lower cut-off point for detection confidence score is used to filter out unreliable detections. In fact, the networks used for the evaluation have the confidence cut-off set at 0.3. Ideally, one wants to deal with the full range of scores, but this is not always realistic without retraining the networks, which is what one wants to avoid in the overall method. 
In all plots in Figure~\ref{fig:moLRP}, moLRP values are omitted where the optimal score threshold $s$ is equal to 0.3. The fact that oLRP finds the optimal score threshold equal to the lower cut-off point means that the true value lies below this level. This can be directly interpreted as relatively poor and unreliable performance. This can be seen as a side-effect of LRP's sensitivity to confidence score (unlike AP). 

When analyzing the detectors' performance, one can observe that as expected, the overall best performance is achieved for H.265 followed by VP9 and H.264. The performance deteriorates below the bitrate of 2.5Mbps, which is clearly visible for SSD networks. An interesting observation is that when it comes to localization performance ($moLRP_{IoU}$) SSD Resnet 50 v1 is the second best for high bitrates (for our dataset). This was not visible using the AP and AR metrics, as seen if Figure~\ref{fig:ap_ar_all_lp}. [mAP@.75 or mAP@[0.5:0.95] ]

Comparing the low-proposal versions of the Faster R-CNN networks (bottom of Figure~\ref{fig:moLRP}), one again sees similar performance with respect to the three LRP components as well as the overall value. Therefore, in most cases, this does not warrant the use of the full version, especially considering the execution time penalties, as shown in the next section.

%

\subsubsection*{Execution Time Evaluation}\label{sec:eval_timing}

When off-loading or complementing computationally intensive tasks such as object detection to a remote unit, the introduction of delays is inevitable. The right side of Figure~\ref{fig:local_vs_remote} presents a schematic of steps involved in processing the image data locally (e.g. directly on a robotic system) as well as using a remote entity. Comparing local processing (in green) with processing on a remote entity, a number of additional steps (in blue) introduce delays in the processing pipeline. These are: encoding and decoding times as well as transmission to and from the remote unit. For off-board processing to be justified, one or both of the following has to be true: (a) the total time of remote processing has to be smaller (due to the availability of more powerful hardware), or (b) it has to provide a solution which is not achievable on the local system alone. In the evaluation, we omit the time taken to grab images, since it is the same for both cases. 

It is important to note that both cases are hardware-dependent, and the reproducibility of the results is not straightforward. The execution speed is highly dependent on the specific hardware configuration and even software versions used. Consequently, the results should be treated as indicators or relative measurements rather than absolute values. Furthermore, in many cases, the inference setup can be tailored to specific hardware to achieve maximum performance. Here, we analyze the performance \emph{as-is} without attempting to improve it through an optimized deployment. At this stage of the investigation, we do not do any comparative evaluation of technologies that can be used to transmit video data between systems. WiFi, 3G, 4G, or even 5G have different properties and their use is dependent on availability. Instead, we provide the time gains of processing using different hardware components. Based on this data, an appropriate transmission technique can be chosen.

Table~\ref{tab:hw_configs} presents the hardware used for the timing evaluation of the models presented in previous sections. Four system configurations, denoted as A, B, C, and D, were used, each with increasing discrete GPU power available. System A is a small form-factor computer from the Intel NUC family. Due to its size, it is suitable for usage on-board in UAV platforms, but it lacks a dedicated discrete GPU. System B is a typical desktop computer with a mid-range gaming-class GPU and a corresponding CPU. System C has been configured for the purpose of machine learning. It is equipped with a GPU tailored for this purpose. System D is a machine in a datacenter. Its configuration is heavily tailored to machine learning tasks. 
\begin{table}[b]
\begin{center}
{
\resizebox{\columnwidth}{!}{%
\begin{tabular}{| c | c| c| c| c | }
\hline
\rowcolor{Gray}
&\textbf{A}&\textbf{B}&\textbf{C}&\textbf{D}\\
\hline
CPU&\makecell{Intel Core i7-7567U, 3.5GHz \\ (max: 4.0GHz), 2+2 cores}&\makecell{ Intel Core i7-6700T, 2.8GHz \\ (max: 3.6GHz), 4+4 cores}&\makecell{Intel Xeon E5-1620 v4, 3.5GHz  \\ (max: 3.8GHz), 4+4 cores}&\makecell{Intel Core (skylake) \\3.2GHz, 4 cores }\\
\hline
\rowcolor{Gray}
GPU&-&Nvidia Geforce GTX 1070, 8GB&Nvidia Titan Xp, 12GB&Nvidia Tesla V100-sXm2, 16GB\\
\hline
RAM&16GB&16GB&16GB&28GB\\
\hline
\rowcolor{Gray}
Location&robot&desktop&desktop&datacenter\\
\hline
\end{tabular}%
}
} 
\caption{Systems used for timing analysis.}
\label{tab:hw_configs}
\end{center}

\end{table}


Table~\ref{tab:time_numbers} presents the average per-frame execution times of the listed networks when performing inference on the evaluation sequences described in Section~\ref{sec:dataset}. Additionally, time differences are listed to easily assess the performance gains in computations using different hardware components. The overall speedup of computations using GPUs compared to CPUs is provided. 

\begin{figure}
\begin{center}
\centerline{\includegraphics[width=0.95\columnwidth]{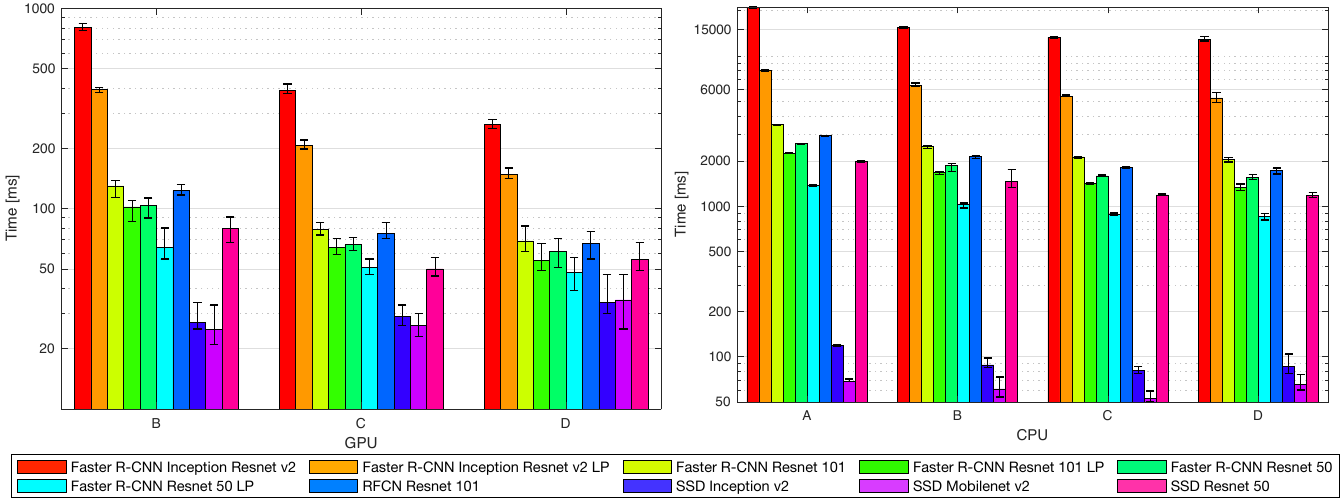}}
   \caption{Inference times (log) for a number of models using three GPU systems. Average, minimum and maximum times are shown.}
   \label{fig:time_gpu_cpu}
\end{center}
\end{figure}

Figure~\ref{fig:time_gpu_cpu} (left), compares the inference performance using GPUs. Average frame inference times as well as minimum and maximum values are provided (note the logarithmic time scale). 
\begin{table}[h]
 \centering
 \resizebox{\columnwidth}{!}{%
 \begin{tabular}{| c | l | r | r | r | r | r | r | r |r | r | r |r | r | r |}
 \cline{3-15}
 
  \multicolumn{2}{c}{} & \multicolumn{4}{|c|}{\textbf{\makecell{CPU\\ {[ms]}}}} & \multicolumn{3}{c|}{\textbf{\textbf{\makecell{GPU\\ {[ms]}}}}}& \multicolumn{3}{c|}{\textbf{\textbf{\makecell{CPU-GPU time diff.\\ {[ms]}}}}}& \multicolumn{3}{c|}{\textbf{\makecell{CPU/GPU\\ speedup}}}\\ 
\cline{1-15}
\multicolumn{2}{|c|}{\textbf{Network}} & \multicolumn{1}{c|}{\textbf{A}} & \multicolumn{1}{c|}{\textbf{B}} & \multicolumn{1}{c|}{\textbf{C}} & \multicolumn{1}{c|}{\textbf{D}}& \multicolumn{1}{c|}{\textbf{B}} & \multicolumn{1}{c|}{\textbf{C}} & \multicolumn{1}{c|}{\textbf{D}}& \multicolumn{1}{c|}{\textbf{B}} & \multicolumn{1}{c|}{\textbf{C}} & \multicolumn{1}{c|}{\textbf{D}}& \multicolumn{1}{c|}{\textbf{B}} & \multicolumn{1}{c|}{\textbf{C}} & \multicolumn{1}{c|}{\textbf{D}}\\
 \hline

\parbox[t]{2mm}{\multirow{6}{*}{\rotatebox[origin=c]{90}{Faster R-CNN}}} 
&\chl Inception Resnet v2&\chl 21030&\chl 15487&\chl 13299&\chl 12969&\chl 801&\chl 390&\chl 263&\chl 14686&\chl 12909&\chl 12706&\chl 19,3&\chl 34,1&\chl 49,3\\ 
&Inception Resnet v2 LP&8012&6412&5403&5270&395&207&149&6017&5196&5121&16,2&26,1&35,4\\ 
&\chl Resnet 101&\chl 3493&\chl 2509&\chl 2130&\chl 2057&\chl 130&\chl 79&\chl 69&\chl 2379&\chl 2051&\chl 1988&\chl 19,3&\chl 27,0&\chl 29,8\\ 
&Resnet 101 LP&2259&1673&1418&1344&102&64&55&1571&1354&1289&16,4&22,2&24,4\\ 
&\chl Resnet 50&\chl 2610&\chl 1860&\chl 1594&\chl 1567&\chl 104&\chl 66&\chl 61&\chl 1756&\chl 1528&\chl 1506&\chl 17,9&\chl 24,2&\chl 25,7\\ 
&Resnet 50 LP&1372&1025&882&860&64&51&48&961&831&812&16,0&17,3&17,9\\ 
\hline
\rowcolor{Gray}
&RFCN Resnet 101&2956&2157&1813&1725&123&75&67&2034&1738&1658&17,5&24,2&25,7\\ 
\hline
\parbox[t]{2mm}{\multirow{3}{*}{\rotatebox[origin=c]{90}{SSD}}}
&Inception v2&118&88&81&86&27&29&34&61&52&52&3,3&2,8&2,5\\ 

&\chl Mobilenet v2&\chl 68&\chl 61&\chl 53&\chl 65&\chl 25&\chl 26&\chl 35&\chl 36&\chl 27&\chl 30&\chl 2,4&\chl 2,0&\chl 1,9\\ 
&Resnet 50 v1&1990&1461&1192&1186&80&50&56&1381&1142&1130&18,3&23,8&21,2\\ 
 \hline
 \end{tabular}%
}
 \caption{Execution times, gains of time and speedup of GPU over CPU processing.}
 \label{tab:time_numbers}
 \end{table}


The gains in performance using the faster GPU systems were as follows: System C was, on average, 1.52 times faster than System B. System D was 1.72 and 1.09 times faster than System B and C, respectively. Unsurprisingly, the most computationally demanding (and most accurate) is the Faster R-CNN ResNet Inception v2. It executes 3.05 and 2.05 times faster using System D, as compared to systems B and C, respectively. In the case of the least computationally demanding models, SSD Inception v2 and SSD Mobilenet v2, even a slight decrease in performance was observed using the more powerful GPU systems C and D. This can be explained by the fact that the more powerful GPUs are not required as these networks have very little demand and the potential gains disappear in the overhead before a GPU can actually process a frame. It is also important to note that these two network configurations make scale input images to a lower resolution of $300\times300$ as specified in the table in Figure~\ref{fig:net_configs}.

The speedup of GPU frame inference due to using a lower number of proposals (as described in Section~\ref{sec:net_configs}) is substantial. For the three variants of networks as ordered in Figure~\ref{fig:time_gpu_cpu} (left), the average speedup for the three GPUs was 1.89, 1.25, and 1.40 times, respectively. 

Figure~\ref{fig:time_gpu_cpu} (right), presents the timing analysis results of performing inference using CPUs. The average, as well as minimum and maximum times are shown (note the logarithmic scale). The differences between performing inference using CPU computations, using the four hardware configurations, are not as large as in the case of GPUs. This is expected, as performing inference using CPUs is known not to be optimal and the CPUs used are similar in computational power. SSD Inception v2 and SSD Mobilenet v2 models are the fastest to compute. These models are the most suitable for use when GPU power is not available, as is often the case with on-board computation on aerial vehicles. 

However, as expected, the gain of using GPUs over CPUs is dramatic. For example, comparing the execution times for the most accurate network, the Faster R-CNN ResNet Inception v2, the gain of the slowest CPU execution (System A) versus the fastest GPU execution (System D) is 80 times (21030ms vs 263ms). This clearly demonstrates the gains of taking advantage of remote computational resources in the Cloud Robotics setting. The speedup gained makes up for any reasonable delays introduced by encoding and transmitting the data for off-board processing. The same applies to the other networks (excluding SSD Inception v2 and SSD 
Mobilenet v2) as the average speedup is 20 times. 

The speedup of frame inference using CPUs is also substantial due to using a lower number of proposals (as described in Section~\ref{sec:networks}). For the three variants of networks as ordered in Figure~\ref{fig:time_gpu_cpu}, the average speedup for the four CPU systems was: 2.49, 1.52, and 1.84 times, respectively. 

The time gains of using GPU over CPU processing (as shown in Table~\ref{tab:time_numbers}) allow us to assess if processing on a remote unit makes sense. For most networks, the time difference is larger than 1 second, which should compensate for the time of encoding, transmitting, and decoding of the video stream. As mentioned earlier, for the most extreme case, the gain in processing time is larger than 20 seconds per frame.

\section{Detector selection}\label{sec:det_choice}

This section considers the problem statement objective for the selection and use of object detector inference setups and algorithms, in addition to the problem statement setup that generically describes the functionality and architectural components assumed. A detector assignment algorithm is specified together with an integer linear programming component that computes the actual allocation of object detectors. The detector assignment algorithm is also evaluated using datasets considered previously in the paper.

\subsection{Problem statement objective}

The paper targets the use of teams of UAVs with ground operation centers for search and rescue missions where the identification of injured humans on the ground is the main priority. Given a team of UAVs, it is assumed that they collectively generate video streams over a target area and can transfer video streams among themselves or to ground operation centers. 

The objective is to process all frames of all video streams generated and maximize the overall accuracy of object detection, in this case, humans on the ground. This implies executing the most accurate detectors in as close to real-time as the physical constraints and mission context allow for. It should be noted that although the focus is on teams of UAVs in search and rescue operations, the generic setup, algorithms, and approach apply to any multi-agent robotic scenario where image processing and object detection are involved.

\subsection{Problem statement setup}

\newcommand{\cplc}{cp_{lcm}}
\newcommand{\cplcm}{$cp_{lcm}$} 

\newcommand{\fpsm}{$fps_{p}$}
\newcommand{\fpst}{processingFrameRate}
\newcommand{\fpstt}{$\fpst (\fps)$}

\newcommand{\dps}{dps}
\newcommand{\dpst}{detPerStream}
\newcommand{\dpstt}{\dpst (\dps)}

\newcommand{\mftt}{maxFrameTime}
\newcommand{\mptt}{maxProcessingTime}

\newcommand{\dpf}{dpf}
\newcommand{\dpft}{detPerFrame}
\newcommand{\dpftt}{\dpft (\dpf)}

\begin{figure}[htb]
   \centering
   \includegraphics[width=0.70\columnwidth]{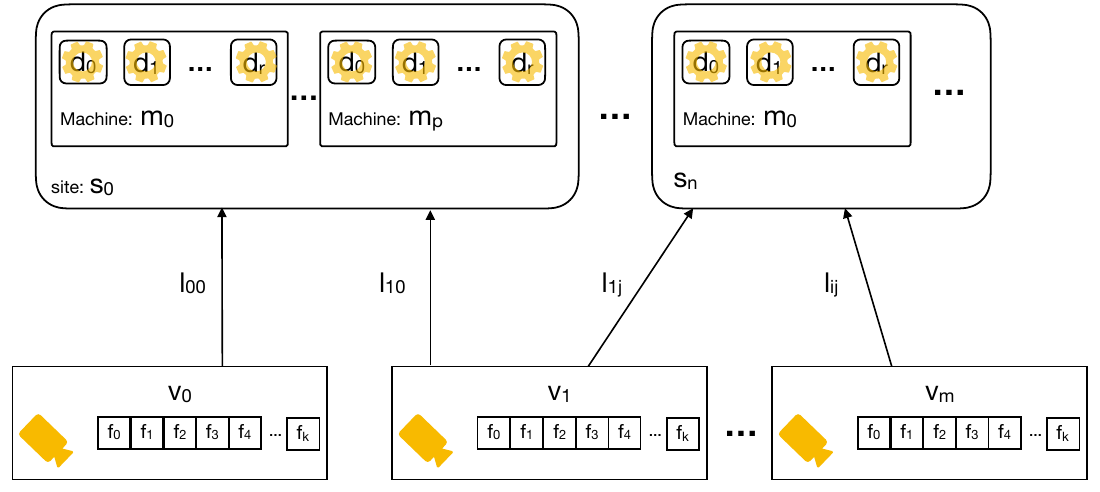}
   \protect\caption{Schematic view of the detector allocation problem. Video streams should be processed using the available detectors running on specific machines allocated to a number of processing sites.}
   \protect\label{fig:detector_allocation}
\end{figure}

The problem statement setup is schematically presented in Figure~\ref{fig:detector_allocation}. Each robotic platform is equipped with one or more camera sensors and the means of transferring video streams generated for additional processing using available communication links. The goal is to process the video streams in as close to real-time as possible at the available processing sites using the available inference setups associated with object detectors. A \emph{processing site} is a location equipped with hardware capable of executing object detection algorithms. It can be the robotic platform itself (on-board processing), a ground station computer, or a machine in a datacenter. The objective is to find the best allocation of image-based detectors so that all video streams are processed with the maximum accuracy possible within the given constraints of a mission.

The starting point is a set of video streams $V=\{v_{0},..,v_{m}\}$ (Figure~\ref{fig:detector_allocation}, bottom) each comprised of frames $F=\{f_{0},..,f_{k}\}$ in each stream which have to be processed. The video contained in the streams has to be transported to one or more of the processing sites $S=\{s_{0},..,s_{n}\}$. At each site $s\in S$, there exists a set of machines $M=\{m_{0},..,m_{p}\}$, which can be physical computers or virtual machines. Each machine can execute a set of vision-based detection algorithms $D=\{d_{0},..,d_{r}\}$ using CPU and or GPU processing. In practice, if a machine is equipped with a GPU, it is preferred over CPU processing. Therefore, without the loss of generality\footnote{A machine equipped with a GPU can be seen by the algorithm as two machines where CPU and GPU processing is done in parallel. } it is assumed that GPU processing takes priority and is performed if it is available. Each aforementioned detector is characterized by the nominal execution time per frame $t$, and the relative $accuracy$ for CPU or GPU processing for a given bitrate (c.f. Table~\ref{tab:time_numbers}). Additionally, each machine is equipped with a certain amount of system and GPU RAM, which allows for the execution of multiple detectors in parallel, as discussed below. Therefore, each detector is also characterized by the amount of memory required for execution.

Transportation of the video streams is achieved using communication links $l_{ij}$ connecting a video source $v_{i}$ to a site $s_{j}$. The main parameter of a link considered here is the available bandwidth:
\begin{equation*}\label{eq:bandwidth}
link\_bw_{i,j}=
\begin{cases}
  0 & \text{if no connection is available},\\
  value & \text{actual maximum bandwidth available},\\
  \infty & \text{within a processing site}.
\end{cases} 
\end{equation*}

Transfer of the video data within a site is assumed to be done without transcoding - a reasonably high bandwidth is available in the form of, for example, wired connections. When transferring a video stream to a site, data is encoded at a specific bitrate $B=\{b_{0},..,b_{s}\}$, for example: \{500kbps, 2Mbps,  5Mbps\}. Without loss of generality,\footnote{The number of encoding bitrate levels can be multiplied by the number of available codecs.} it is assumed that only one encoding technique is used for all transferred streams and communication links. 

As stated above, the objective is to process all frames of all video streams in $V$ and maximize the overall detection accuracy, which means executing the most accurate detectors in as close to real-time as the physical constraints allow for. The physical constraints are the available bandwidth of links, the number of sites, the number of machines at each site, and the number of available detectors with their execution requirements.

The following subsections describe an algorithm for solving the problem objectives stated above. The solution makes use of an Integer Linear Programming (ILP) approach for optimally assigning detectors for each frame of all the video streams to be processed.

\subsubsection*{Detector assignment}

The detector assignment algorithm is presented in Algorithm~\ref{alg:det_ass}. It consists of a data pre-processing stage and the execution of an ILP solver. The input data for the algorithm consists of several parameters. The main input parameter is the $problemSetup$, which consists of a set of video streams to be processed $V$, the available processing sites $S$, properties of machines $M$ at each site (including the available detectors $D$), and the parameters of the communication links. 

\IncMargin{1em}
\begin{algorithm}[h]
\DontPrintSemicolon
\SetAlgoLined
\SetNoFillComment 

\SetKwData{setup}{problemSetup}
\SetKwData{maxprocessingtime}{\mptt}
\SetKwData{maxframetime}{\mftt}
\SetKwData{wantedframerate}{processingFrameRate}
\SetKwData{detperstream}{detPerStream}
\SetKwData{detperframe}{detPerFrame}
\SetKwData{lcm}{\cplcm}
\SetKwData{detectors}{detectors}
\SetKwData{ass}{assignment}

\SetKwFunction{getlcm}{getLCM}
\SetKwFunction{solveilp}{solveILP}
\SetKwFunction{getdet}{getDetectors}

\SetKwInOut{input}{Input}
\SetKwInOut{output}{Output}

\input{$\setup$, $\wantedframerate$, $\maxprocessingtime$, $\detperstream$, $\detperframe$}  

\output{detector assignment: \ass}

\BlankLine
$\ass \leftarrow \emptyset$

$\maxframetime \leftarrow min(\maxprocessingtime, \detperstream*1/\wantedframerate)$\label{ass:line:maxframetime}

\tcc{get the list of detectors with compatible execution time}
$\detectors \leftarrow \getdet(\setup, \maxframetime)$ \label{ass:line:use_maxframetime}

\If{$\detectors $ == $ \emptyset$}{
	return $\ass$
}

\tcc{compute least-common-multiple}
$\lcm \leftarrow \getlcm(\detectors)$ \label{ass:line:get_lcm}

\tcc{solving ILP problem}
$\ass \leftarrow \solveilp(\setup, \lcm)$ \label{ass:line:ilp}

return \ass\;\label{ass:line:return}
\caption{Detector assignment algorithm.}\label{alg:det_ass}
\end{algorithm}
\DecMargin{1em}

The parameter $processingFrameRate$ defines at what framerate the processing should be carried out, that is, how many frames need to have a detector assigned. The value of this parameter depends on the framerates of the cameras used. Additionally, depending on the mission at hand, it might not be required to process the video streams at the full framerates. If a lower framerate is chosen, potentially more accurate detectors can be assigned due to more frame processing time available. The second parameter, $\mptt$, specifies how long a time is allowed between receiving a frame of video until the result is available. This parameter decides which of the available detectors can be considered when assigning, as the execution time of a detector has to be smaller or equal to $\mptt$.

\begin{figure}[htb]
   \centering
   \includegraphics[width=1.00\columnwidth]{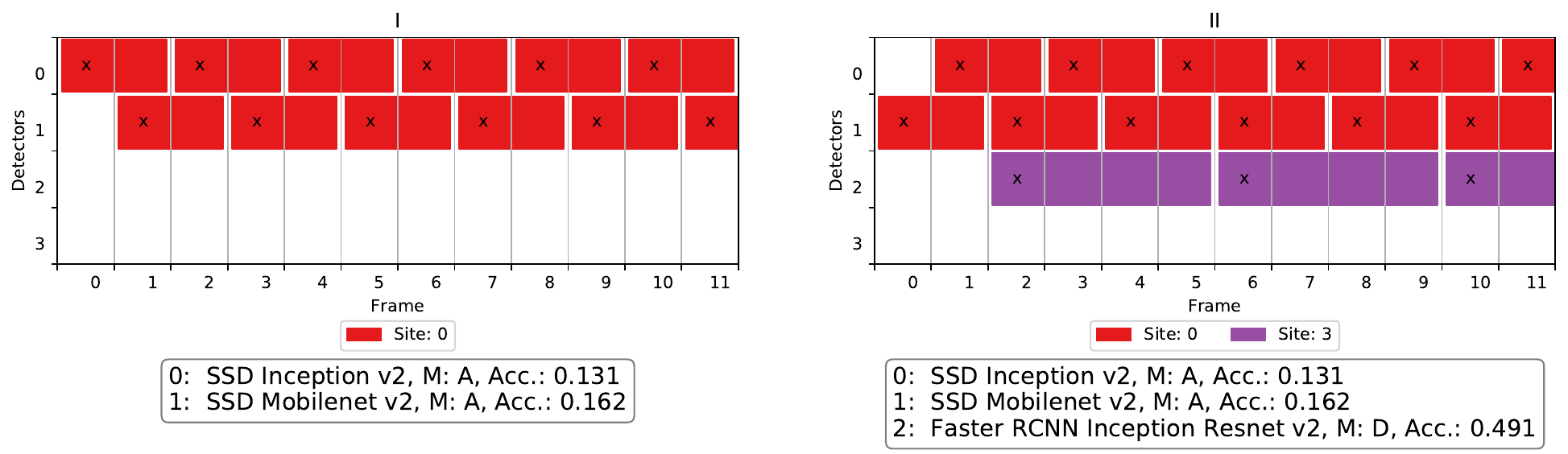}
   \protect\caption{Two example detector allocations for $\dpst$ = 4, $\dpft$ = 1 (left), $\dpft$ = 2 (right). Maker 'x' denotes the frame to be processed and the execution time is reflected by a rectangle.}
   \protect\label{fig:dpf}
\end{figure}

Parameters $\dpst$ and $\dpft$ allow for specifying the maximum number of detectors which are allowed to be used to process each stream (the former) and each frame (the latter). Figure~\ref{fig:dpf} shows example assignments for a single stream with $\dpst$ = 4 (therefore four rows in the figure). It presents two allocations of the 12 first frames of a cyclic schedule. On the left, each frame is allowed to be processed only once ($\dpft$=1). The optimal schedule found consists of two detectors (listed in a box below the plot). On the right, two detectors are allowed to process one frame ($\dpft$=2) which results in frames 2, 6, and 10  processed by an additional detector. In both cases, even though four detectors are allowed to be used per video stream, the optimum requires only 2 or 3 detectors to be allocated.  

Algorithm~\ref{alg:det_ass} works as follows. First, in line~\ref{ass:line:maxframetime}, the value of $\mftt$ is calculated as a minimum between $\mptt$ and a time dictated by the number of detectors allowed to be used per stream ($\dpst$). For example, if $\dpst$=1, only detectors with periods smaller than the inverse of the processing frame rate should be considered, etc. The following step of the algorithm (line~\ref{ass:line:use_maxframetime}) is used to obtain a set of available detectors $D$  taking into account the maximum execution time on their respective hardware. Directly after, the least-common-multiple of periods of the detectors is calculated (\cplcm, line~\ref{ass:line:get_lcm}). The interest is in assigning detectors to process frames that is analogous to obtaining a non-preemptive cyclic schedule where detectors are jobs with a specific execution time. If a \cplcm is equal to C, the schedule is cyclic after C amount of time units. The role of the following step of the algorithm (line~\ref{ass:line:ilp}) is to assign detectors to process the available video streams, that is, to find a cyclic schedule that respects all the constraints described in the following subsection. Finally, the results are returned in line~\ref{ass:line:return}.

\subsection{Integer programming formulation}

The actual allocation of detectors is achieved using an Integer Linear Programming (ILP) technique (Algorithm~\ref{alg:det_ass}, line \ref{ass:line:ilp}). The variables of the ILP formulation are presented in Table~\ref{tab:ilp_variables}. The binary variable $x$ denotes whether a specific detector $d$ is assigned to process video stream $v$ encoded at bitrate $d$. The binary variable $y$ specifies if a frame $f$ of video stream $v$ is processed by a detector $d$. Although it is similar to the $x$ variable, it introduces the required level of granularity necessary for checking constraints in Equations~\ref{eq:ilp_constr_detperframe} and~\ref{eq:ilp_frame_timing} described below, which deal with specific frames of video streams. 

\begin{table}[h]
\begin{center}
{
\begin{tabular}{ l | l  }
\textbf{Symbol}&\textbf{Meaning}\\
\hline
$x_{vdb} \in \{0,1\}$& 1 if a detector $d$ is used to process video stream $v$ at bitrate $b$, 0 otherwise \\
$y_{vdf} \in \{0,1\}$& 1 if a frame $f $ of video stream $v$ is processed using detector $d$, 0 otherwise \\
$linkUsed_{vs}\in\{{0,..,\vert S \vert}\}$& Number of times a link is used to transport a stream $v$ to site $s$\\ 
\end{tabular}
} 
\caption{Variables used in the problem.}
\label{tab:ilp_variables}
\end{center}

\end{table}

The parameters used in the problem specification and the algorithm are presented in Table~\ref{tab:ilp_params}.

\begin{table}[h] 
\begin{center}
{
\begin{tabular}{ l | l  }

\textbf{Symbol}&\textbf{Meaning}\\
\hline
$V$& Set of video streams to be processed\\
$S$& Set of processing sites \\
$M$& Set of machines\\
$D$& Set of all the available detectors compatible with $maxFrameTime$ (from all sites and machines)\\
$B$& Set of the considered bitrate levels compatible with $max\_link\_bw$ \\
$ram_{m}$&Amount of RAM memory of the machine or the GPU $m\in M$\\
$ram_{d}$& Required amount of RAM for detector $d\in D$ \\
$period_{d}$& Period (i.e. 1/execution time) of a detector $d\in D$ \\
$acc_{d}^{b}$& Accuracy of a detector $d\in D$ at bitrate $b\in B$\\
$bitrate_{b}$& Specific bitrate level $b$ such that $\sum_{b}^{\vert B\vert} bitrate_{b} = B$\\
$link\_bw_{v}^{s}$& Cost of transferring of a video stream $v\in V$ to a processing site $s\in S$\\
$max\_link\_bw$ & Maximum link bandwidth of $link\_bw_{v}^{s}$\\
\hline
$D_{m}$ & Detectors of a given machine, i.e. subset of $D$ such that $\sum_{m}^{\vert M\vert} D_{m} = D$\\ 
$D_{s}$ & Detectors at a given site, i.e. subset of $D$ such that $\sum_{s}^{\vert S\vert} D_{s} = D$\\ 
\cplcm& Cyclic period calculated as least-common-multiple of detectors' periods.\\
\end{tabular}

} 
\caption{The problem input parameters and the data resulting from the pre-processing (bottom part).}
\label{tab:ilp_params}
\end{center}
\end{table}

The objective of the ILP function takes the following form:
\begin{equation}
acc_{obj}=\sum_{v\in V}\sum_{d\in D}\sum_{b\in B} \frac{bitrate_{b}}{period_{d}} *acc_{d}^{b}*x_{vdb} 
\protect\label{eq:ilp_objective1}
\end{equation}
For all video streams and each detector used at a specific bitrate ($x_{vdb}$), it sums up the accuracy of the detector at a specific bitrate ($acc_{d}^{b}$), promotes higher bitrates ($bitrate_{b}$), and takes into account how many frames are processed by the specific detector ($period_{d}$). 

In order to minimize the number of sites used, the following penalty term is used: 
\begin{equation}\protect\label{eq:ilp_penalty}
penalty=\sum_{v\in V}^{} \sum_{s\in S} linkUsed_{vs} 
\end{equation}

The penalty term is the number of processing sites used calculated as a sum of values obtained using the constraint in Equation~\ref{eq:ilp_contr_link}.

The ILP problem is defined as follows:\\
\textbf{Maximize:}

\begin{equation}\protect\label{eq:objective}
acc_{obj}*w - penalty*(1-w)\\
\end{equation}
 
w.r.t.:

\begin{equation}\protect\label{eq:ilp_constr_detperstream}
\sum_{b\in B}^{} \sum_{d\in D}^{} x_{vdb} \leq \dpst  \mathbigspace \forall v \in V
\end{equation}
\begin{equation}\protect\label{eq:ilp_constr_detperframe}
1 \leq\sum_{d\in D}^{} y_{vdf} \leq detPerFrame \mathbigspace \forall v \in V, f \in F
\end{equation}
\begin{equation} \protect\label{eq:ilp_frame_timing}
\sum_{i=k}^{k+period_{d}} y_{vdf} = \sum_{b\in B}^{} x_{vdb} \mathbigspace \forall k\in\{0,... ,\cplc-period_{d}+1\}, v \in V, d\in D
\end{equation}
\begin{equation}\protect\label{eq:ilp_contr_bw}
\sum_{d\in D_{s}}^{} \sum_{b\in B}^{} \frac{bitrate_{b}}{period_{d}}* x_{vdb} \leq linkBw_{v}^{s}  \mathbigspace \forall v\in V, s \in S
\end{equation}
\begin{equation}\protect\label{eq:ilp_contr_ram}
\sum_{v\in V}^{} \sum_{d\in D_{m}}^{} \sum_{b\in B}^{} ram_{d} * x_{vdb} \leq ram_{m} \mathbigspace \forall m\in M 
\end{equation}
\begin{equation}\protect\label{eq:ilp_contr_link}
\sum_{d\in D_{s}}^{} \sum_{b \in B}^{} x_{vdb} \leq M * linkUsed_{vs} \mathbigspace \forall v\in V, \forall s \in S 
\end{equation}

The objective based on accuracy in Equation~\ref{eq:ilp_objective1} and the penalty term in Equation~\ref{eq:ilp_penalty} are weighted by a constant $w$. This allows for balancing which property of the solution should dominate: the accuracy at the cost of video transport or vice versa. For example, for $w$ close to 1, little consideration is given to the number of links used, while for $w$ close to 0 the amount of links used will be minimized (notice that the maximum allowed bitrate will still be used). 

The constraint expressed in Equation~\ref{eq:ilp_constr_detperstream} ensures that each stream is processed by the required number of detectors: $\dpst$. Equation~\ref{eq:ilp_constr_detperframe} ensures that for each video stream, each frame is processed by the required number of detectors: at least one and not more than $detPerFrame$. 
The constraint expressed in Equation~\ref{eq:ilp_frame_timing} ensures that assigning detectors to frames is performed without violating the execution periods of the selected detectors. This is achieved by a sliding window approach i.e. the sum of frames within a window of length equal to the detectors period is 1 if the detector is assigned to be used (i.e. the meaning of the $x$ variable), and 0 otherwise. This constraint also ensures that only one bitrate level $b$ is used for each detector. Equation~\ref{eq:ilp_contr_bw} ensures that if a specific encoding bitrate is used with a detector processing a video stream, it is compatible with the physical bandwidth available connecting the stream with the processing site. Equation~\ref{eq:ilp_contr_ram} ensures that the RAM needed by all detectors executed on a specific machine is less than the available amount of memory of that machine. 

The constraint in Equation~\ref{eq:ilp_contr_link} iterates through the sites and video stream and counts the number of times a link is used to transport the video at one of the available bitrates. The constraint uses a constant M which is equal to the maximum number of sites (i.e. $\vert S\vert$). The result is that $linkUsed_{vs}$ is equal to 1 if at least one video is transported between $v$ and $s$, 0 otherwise. This information is used by the aforementioned penalty term of the objective function (Equation~\ref{eq:ilp_penalty}). 



\subsection{Evaluation}
Integer Linear Programming is an NP-hard problem. The space complexity (i.e. number of variables) of our formulation can be calculated using the following formula: 
\begin{equation}
\textbf{number of variables}: \vert V\vert *(\vert D \vert *(\vert B\vert+\vert F\vert) +\vert S\vert)
\protect\label{eq:number_of_vars}
\end{equation}
It grows with the number of videos to be processed~$(V)$, the number of the available detectors~$(D)$, the number of compression levels $(B)$ considered, the common period for detectors~$(F, \vert F\vert = \cplc)$, and finally, the number of processing sites $(S)$. The number of constraints for a specific instance of the problem can be calculated as:
\begin{equation}
\begin{split}
\textbf{number of constraints}:  \vert M \vert + \vert V \vert *(1+ 2(\vert F \vert + \vert S\vert) ) + \\
({\vert F \vert-period_{d}+1}) * \vert V \vert  \;  \forall d \in D
\end{split}
\protect\label{eq:number_of_contr}
\end{equation}
It depends on the number of processing sites (S), machines (M), video streams to be processed (V), detectors (D) and the common period for detectors (F). Additionally, it depends on specific periods of detectors (\textit{$period_d$}) as evident from the second line of the constraint equation~\ref{eq:ilp_frame_timing}).

When applying ILP in the detector assignment algorithm, the values of variables often have practical limits. For example, when dealing with in-the-field operation in a setting such as Search and Rescue, a major limiting factor is the bandwidth of communication links available for transmitting video signals to processing sites such as datacenters. For this reason, the size of $V$ as well as $B$ are limited in practice. Similarly, the number of available machines that can perform computations can get prohibitively expensive mainly due to the relatively high prices of GPUs, which are preferred for many convolutional network-based algorithms. For these reasons, the algorithm was evaluated with the following upper bound values: up to 150 video streams at 20Mbps between a stream and a remote location and up to 40 processing units available for use. These values provide a generous upper bound on realistic operational scenarios for search and rescue.




The influence of the aforementioned parameters have been evaluated using the data obtained during the detector evaluation, presented in Section~\ref{sec:networks}. The detector period information was derived from the worst-case execution timing information presented in Figure~\ref{fig:time_gpu_cpu}, using the following equation:
\begin{equation}\protect\label{eq:period}
period_{d}=\ceil*{time_{d}*\fpst}
\end{equation}
Table~\ref{tab:ilp_periods} presents the periods of the detectors calculated for a processing rate of 30Hz. The highlighted cells show the periods compatible with the $\dpst$ from 1 to 5 used during the evaluation.

For the case of execution of detectors in parallel on a single machine,  the same execution timing for these detectors was assumed. In practice, a separate evaluation of the influence of parallel execution would have to be performed and this is relegated to future analysis. This assumption, however, does not influence the performance of the algorithm. Alternatively, it is possible to replace the constraint in Equation~\ref{eq:ilp_contr_ram} with the following one:
\begin{equation}
\sum_{v\in V}^{} \sum_{d\in D_{m}}^{} \sum_{b\in B}^{} x_{vdb} \leq 1 \mathbigspace \forall m\in M 
\protect\label{eq:ilp_contr_ram_simpl}
\end{equation}
This simplified constraint ensures that a machine can be used only once. Using this constraint would improve the timing performance of the algorithm as it is more restrictive. 
\begin{table}[h]
 \centering
 \begin{tabular}{|c|l|r|r|r|r|r|r|r|}
 \cline{3-9}
   \multicolumn{2}{c}{} & \multicolumn{7}{|c|}{\textbf{System}} \\ 
 \cline{3-9}
  \multicolumn{2}{c}{} & \multicolumn{4}{|c|}{\textbf{CPU}} & \multicolumn{3}{c|}{\textbf{GPU}}\\ \cline{1-9}
  \multicolumn{2}{|c|}{\textbf{Network}} & \multicolumn{1}{c|}{\textbf{A}} & \multicolumn{1}{c|}{\textbf{B}} & \multicolumn{1}{c|}{\textbf{C}} & \multicolumn{1}{c|}{\textbf{D}}& \multicolumn{1}{c|}{\textbf{B}} & \multicolumn{1}{c|}{\textbf{C}} & \multicolumn{1}{c|}{\textbf{D}}\\
 \hline

\parbox[t]{2mm}{\multirow{6}{*}{\rotatebox[origin=c]{90}{Faster R-CNN}}} 
&Inception Resnet v2&316&233&200&195&13&6&\chl4\\ 
&Inception Resnet v2 LP&121&97&82&80&6&\chl4&\chl3\\ 
&Resnet 101&53&38&32&31&\chl2&\chl2&\chl2\\ 
&Resnet 101 LP&34&26&22&21&\chl2&\chl1&\chl1\\ 
&Resnet 50&40&28&24&24&\chl2&\chl1&\chl1\\ 
&Resnet 50 LP&21&16&14&13&\chl1&\chl1&\chl1\\ 
\hline
&RFCN Resnet 101&45&33&28&26&\chl2&\chl2&\chl2\\ 
\hline
\parbox[t]{2mm}{\multirow{3}{*}{\rotatebox[origin=c]{90}{SSD}}}
&Inception v2&\chl2&\chl2&\chl2&\chl2&\chl1&\chl1&\chl1\\ 
&Mobilenet v2&\chl2&\chl1&\chl1&\chl1&\chl1&\chl1&\chl1\\ 
&Resnet 50 v1&30&22&18&18&\chl2&\chl1&\chl1\\ 
 \hline
 \end{tabular}
 \caption{Periods of detectors at $\fpst$=30 used for the algorithm evaluation. The highlighted values conform to the value of $\dpst \leq 5$ . Low proposal versions (LP) of the networks were not taken into consideration. }
 \label{tab:ilp_periods}
 \end{table}
 

The evaluation consisted of measuring computation times\footnote{The implementation uses Google OR-Tools SAT solver (through Python bindings, version: 7.5.7466),  Quad-Core Intel Core i7 2.9Ghz, 6th generation (6920HQ), 16MB of RAM.} for varying numbers of video streams to be processed for 40 machines (10 machines of types A, B, C, and D), where each machine has all the detectors available (subject to $\mptt$ and $\dpst$). Several link bandwidth values between streams and sites were evaluated and representative results for two are presented: 1Mbps and 20Mbps. These values allow for considering 6 and 9 compression bitrate levels, respectively. 


Even though, in general, the timing performance of an ILP algorithm cannot be reliably measured just by using only the number of variables (rows) and constraints (columns), the two quantities calculated using Equations~\ref{eq:number_of_vars} and~\ref{eq:number_of_contr} are shown  below the times (separated by a colon), for each case. 

\begin{table}[h]

\begin{center}
{
\begin{tabular}{ c c c c c  c  c  c c c c  c c c}
\toprule
 &&&&\multicolumn{9}{c }{\textbf{Number of streams} $\vert V\vert$}\\ \cmidrule(lr){5-13}
&$\dps$& $\vert D\vert$&\textbf{\cplcm} & \textbf{1}&\textbf{2}&\textbf{5}&\textbf{10}&\textbf{15}&\textbf{25}&\textbf{30}&\textbf{40}&\textbf{50}\\\hline
&1&40&\textbf{1}&0.1&0.1&0.3&0.7&1.1&1.8&2.1&2.8&3.7\\
&&&&\tiny364:91&\tiny728:142&\tiny2k:295&\tiny4k:550&\tiny5k:805&\tiny9k:1k&\tiny11k:2k&\tiny15k:2k&\tiny18k:3k\\
&2&130&\textbf{2}&0.1&0.3&0.8&2.0&6.6&16.0&24.8&48.4&73.3\\
&&&&\tiny1k:223&\tiny3k:406&\tiny7k:955&\tiny13k:2k&\tiny20k:3k&\tiny33k:5k&\tiny39k:6k&\tiny52k:7k&\tiny65k:9k\\
&3&170&\textbf{6}&0.3&0.7&2.1&6.4&12.0&40.8&56.0&142.4&300.9\\
&&&&\tiny2k:911&\tiny5k:2k&\tiny12k:4k&\tiny24k:9k&\tiny36k:13k&\tiny60k:22k&\tiny72k:26k&\tiny95k:35k&\tiny119k:44k\\
&4&200&\textbf{12}&1.0&1.5&4.8&11.6&26.9&86.5&171.2&>600&>600\\
&&&&\tiny4k:2k&\tiny8k:4k&\tiny20k:11k&\tiny40k:22k&\tiny60k:33k&\tiny100k:54k&\tiny120k:65k&\tiny160k:87k&\tiny200k:109k\\
&5&210&\textbf{60}&4.5&8.6&33.8&169.8&371.6&>600&>600&>600&>600\\
&&&&\tiny14k:12k&\tiny29k:25k&\tiny71k:62k&\tiny143k:124k&\tiny214k:186k&\tiny357k:311k&\tiny429k:373k&\tiny571k:497k&\tiny714k:621k\\
\hline
\multicolumn{13}{c }{(a) For $link\_bw_{v}^{s}$ = 1Mbps}\\
\bottomrule
\end{tabular}

\begin{tabular}{ c c c c c  c  c  c c c c  c c c}
\toprule
&$\dps$& $\vert D\vert$&\textbf{\cplcm} & \textbf{1}&\textbf{2}&\textbf{5}&\textbf{10}&\textbf{15}&\textbf{25}&\textbf{30}&\textbf{40}&\textbf{50}\\\hline

&1&40&\textbf{1}&0.1&0.2&0.4&0.8&1.5&2.2&2.7&3.4&4.5\\
&&&&\tiny404:91&\tiny808:142&\tiny2k:295&\tiny4k:550&\tiny6k:805&\tiny10k:1k&\tiny12k:2k&\tiny16k:2k&\tiny20k:3k\\
&2&130&\textbf{2}&0.1&0.3&0.8&1.8&6.1&15.3&26.2&40.2&82.2\\
&&&&\tiny1k:223&\tiny3k:406&\tiny7k:955&\tiny14k:2k&\tiny22k:3k&\tiny36k:5k&\tiny43k:6k&\tiny57k:7k&\tiny72k:9k\\
&3&170&\textbf{6}&0.3&0.8&2.2&4.1&7.4&21.5&38.2&49.0&115.5\\
&&&&\tiny3k:911&\tiny5k:2k&\tiny13k:4k&\tiny26k:9k&\tiny38k:13k&\tiny64k:22k&\tiny77k:26k&\tiny102k:35k&\tiny128k:44k\\
&4&200&\textbf{12}&0.7&1.5&3.6&12.4&18.8&61.2&164.1&>600&>600\\
&&&&\tiny4k:2k&\tiny8k:4k&\tiny21k:11k&\tiny42k:22k&\tiny63k:33k&\tiny105k:54k&\tiny126k:65k&\tiny168k:87k&\tiny210k:109k\\
&5&210&\textbf{60}&49.2&>600&>600&>600&>600&>600&>600&>600&>600\\
&&&&\tiny14k:12k&\tiny29k:25k&\tiny72k:62k&\tiny145k:124k&\tiny217k:186k&\tiny362k:311k&\tiny435k:373k&\tiny580k:497k&\tiny725k:621k\\
\hline
\multicolumn{13}{c }{(b)  For $link\_bw_{v}^{s}$ = 20Mbps}\\
\bottomrule
\end{tabular}
} 
\caption{Computation times in seconds for allocating the specified number of video streams. Below the execution time: the number of problem variables and constraints separated by a colon. Number of detectors ($\vert D\vert$) and common period (\cplcm) is provided for each value of $\dpst$ ($dps$). Parameters: 40 machines at 4 sites (10 of each type A, B, C, and D), $detPerFrame$=2, $processingFrameRate$=30Hz, $w$=0.6. Top: 6 bitrate level ($\leq$1Mbps) $B_{6} = \{50, 75, 100, 250, 500, 1000\}$kbps, bottom: 9 bitrates: $B_{20}=  B_{6} \cup \{2500, 5000, 20000\}$kbps.} 
\label{tab:ilp_eval1}
\end{center}
\end{table}

The algorithm execution timing results presented in Table~\ref{tab:ilp_eval1} show general trends of the behavior of the algorithm as the main parameters grow, namely the number of video streams to be processed (V) as well as the number of detectors to be used by each video stream (D). The influence of the number of encoding levels ($B$) is not significant. It is mostly visible when combined with larger values of $\dpst$ (e.g. 5). It is not consistent between the two evaluated cases. It is evident by the case for $\dpst$ = 4. The execution time is slightly decreased in the latter case. It might be due to the problem structure or even the properties of the solver. The differences are not big, and the general trend is followed. 

Further results presented in Table~\ref{tab:ilp_eval2}, show the influence of even larger numbers of video streams, but for lower number of detectors allowed to be used per stream i.e. $\dpst \leq2$. In the case of allowing only one detector to be used per video stream, the calculation times stay low. Even when no solution exists (as indicated with a "*" symbol) the algorithm terminates in a timely manner. When two detectors are allowed ($\dpst$=2) the algorithm delivers a result in a reasonable time for up to 75 video streams. 

\begin{table}[h] 
\begin{center}
{
\begin{tabular}{ c c c c c  c  c  c c c }
\toprule
 &&&&\multicolumn{5}{c }{Number of streams $\vert V\vert$}\\ \cmidrule(lr){5-9}
&$dps$& $\vert D\vert$&\textbf{\cplcm} & \textbf{60}&\textbf{75}&\textbf{100}&\textbf{125}&\textbf{150}\\\hline
\bottomrule
&1&40&\textbf{1}&6.4&6.8&9.0&61.5*&48.8*\\
&&&&\tiny24k:3k&\tiny30k:4k&\tiny40k:5k&\tiny50k:6k&\tiny61k:8k\\
&2&130&\textbf{2}&115.6&102.2&>600&>600&>600\\
&&&&\tiny86k:11k&\tiny108k:14k&\tiny143k:18k&\tiny179k:23k&\tiny215k:27k\\
\end{tabular}
} 
\caption{Execution times in seconds of allocating the specified number of video streams to 40 machines (10 of each types A, B, C, and D) using 9 bitrate levels. LinkBW 20. (*) Indicates that no solution exists. Parameters: $detPerFrame$=2, $processingFrameRate$=30Hz. } 
\label{tab:ilp_eval2}
\end{center}
\end{table}

Although in this approach, the selection and allocation of detectors is performed once before a mission starts, the conditions in the field can change, for example, the communication links can become better or worse. In such cases, a new allocation could be performed to take advantage of or adjust to the new conditions. Dynamic change of allocation, however, is also relegated to future work.

The evaluation in this section and experience in the field provide strong evidence that the detector allocation algorithm proposed is robust and pragmatically useful in the field with reasonable soft real-time behavior.




\section{Detection fusion and map building}\label{sec:fusion}

During a Search and Rescue mission, it is assumed that a number of UAS platforms collect video data, which is then processed by specific object detection algorithms previously allocated at the start of a mission. This allocation process has been described in the previous sections. In this section, the process of acquiring and fusing a collection of vision-based detections is considered. The objective is to use these detections to generate saliency maps detailing the geolocations of objects discovered during the detection process.

The proposed approach for fusing vision-based object detections takes advantage of a probabilistic framework laid down in~\cite{hor:13_octomap}, a mapping technique based on octrees. The original method uses probabilistic occupancy estimation based on range sensors such as LiDARs or RGB-D cameras. A voxel occupancy probability is updated taking into account the log-odds relation $L(n)=log\left(\frac{P(n)}{1-P(n)}\right)$ by a simple addition as follows: 
\begin{equation}
L(n\given z_{1:t})= L(n\given z_{1:t-1}) + L(n\given z_{t})
\protect\label{eq:voxelloupdate}
\end{equation}

Instead of using range measurements to update the occupancy, this paper defines the update $L(n\given z_{t})$ using one or more object detections from one or more robotic platforms to add information about occupied locations containing objects of different classes. It is achieved using a sensor model that allows for incorporating detection results (or lack thereof), taking into account the bounding boxes obtained, confidence scores, properties of the detectors used, properties of the detected objects, and the state of the sensors used to collect the raw data. By taking into account all this available data, the model allows for robustly incorporating new information as well as rejecting false positive detections, which are not uncommon even when using accurate detectors, CNN-based or otherwise.

\newcommand{\areafactor}{p_{bbox}^{area}}
\newcommand{\distfactor}{p_{distance}}
\newcommand{\detfactor}{p_{det}^{rel}}
\newcommand{\detscore}{p_{det}^{score}}
\newcommand{\maxnegative}{p_{negative}^{max}}
\newcommand{\maxpositive}{p_{positive}^{max}}
\newcommand{\objareamax}{obj_{w|h}^{max}}
\newcommand{\objareamin}{obj_{w|h}^{min}}
\newcommand{\bbareamax}{bbox_{area}^{max}}
\newcommand{\bbareamin}{bbox_{area}^{min}}
\newcommand{\imsize}{img_{w|h}^{input}}
\newcommand{\objsize}{obj_{x|y}}
\newcommand{\imsizecalib}{img_{w|h}^{calib}}
\newcommand{\camobjdist}{dist_{cam}^{obj}}





\subsection{Map update and salient locations calculation algorithm}

\begin{algorithm}[t]
\DontPrintSemicolon
\SetAlgoLined
\SetNoFillComment 

\SetKwData{Det}{det}
\SetKwData{GeoLoc}{geoloc}
\SetKwData{map}{G}
\newcommand{\gindex}{index}
\SetKwData{index}{index}
\SetKwData{detections}{D}
\SetKwData{salientmap}{S}
\SetKwData{clusters}{C}
\SetKwData{cluster}{c}
\SetKwData{weights}{LOC}
\SetKwData{weight}{w}
\SetKwData{pthresh}{threshold}
\SetKwData{clustparams}{parameters}

\SetKwData{dist}{MAX\_DIST}
\SetKwData{det}{d}
\SetKwData{pose}{pose}
\SetKwData{bin}{bin}
\SetKwData{voxels}{voxels}
\SetKwData{voxel}{v}
\SetKwData{distance}{distance}
\SetKwData{camloc}{loc\_cam}
\SetKwData{pixel}{p}
\SetKwData{pixeldist}{u}
\SetKwData{Image}{image}
\SetKwData{CameraInfo}{CameraInfo}

\SetKwFunction{computeWeight}{computeWeight}

\SetKwInOut{Input}{input}
\SetKwInOut{Output}{output}

\Input{a set of detections: $\detections$, camera pose: $\pose$}  
\Input{clustering parameters: $\clustparams$, $\pthresh$}  
\Output{voxel grid map: \map, list of salient locations: \salientmap}

\BlankLine
\tcc{update based on object detection results (bounding box, score, detector id etc.}
	\ForAll{$\det \in \detections$}{
		$\map \leftarrow positiveUpdate(\map, \det)\{$\; \label{alg1:line:positive}
			\Indp 
			$\voxel \leftarrow getGridVoxel(\map, \pose, \det)$\;
			\uIf{$\voxel$ is valid}{
				$\distance \leftarrow getDistance(\det, \pose, \voxel)$\;
				$p \leftarrow getProbabilityPositive(\det, \distance)$ \tcp*[h]{Equation~\ref{eq:p_positive}}\;\label{alg1:line:positive_update}
				$\map \leftarrow updateProbability(\voxel, p)$\;\label{alg1:line:positive_update_end}
   			} 
		\Indm \}\ \;
		
    	}
    \tcc{update based on the camera pose}

     $\map \leftarrow negativeUpdate(\map, \pose)\{$\; \label{alg1:line:negative}
			\Indp 
			$\voxels \leftarrow getVisibleVoxels(\map, \pose)$\;\label{alg1:line:get_indexes}			
			\ForAll{$\voxel \in \voxels$}{
				$distance \leftarrow getDistance(\voxel, \pose)$\;
				$p \leftarrow getProbabilityNegative(\pose, distance)$ \tcp*[h]{Equation~\ref{eq:p_negative}}\;\label{alg1:line:negative_update}
				$ \map \leftarrow updateProbability(\voxel, p)$\;
   			}			
		\Indm \}\
  
	$S \leftarrow extractSalientLocations(\map)\{$\;\label{alg1:line:get_salient}
			\Indp 
			$G' \leftarrow thresholdGridMap(G, \pthresh )$\;\label{alg1:line:threshold}
            $S \leftarrow getSalientLocations(G', \clustparams)$\;
            
		\Indm \}\
	
return \map, \salientmap
\caption{Grid map update and salient locations calculation}\label{alg:grid_update}
\end{algorithm}

The process of fusing detection results is outlined in Algorithm~\ref{alg:grid_update}. First, for each detection in a given set $D$ (corresponding to a single image frame), a positive update of the grid map is calculated based on the detections obtained (hence the designation: \emph{positive}, lines~\ref{alg1:line:positive}-\ref{alg1:line:positive_update_end}). 
A positive update starts by performing a lookup of a specific voxel corresponding to a detection bounding box. The voxel is valid if a grid cell exists in the map and it is invalid in case of a false positive detection corresponding to a non-existent grid cell. For a valid voxel, a distance between the camera and the grid cell is calculated, which is used to calculate the positive update of the probability (line~\ref{alg1:line:positive_update}, described below).

A negative update of the map starts in line~\ref{alg1:line:get_indexes} by obtaining a set of grid voxels in the camera's field of view, given the camera's pose. Object detection is performed for each image, and it may contain no detections and, therefore, no bounding boxes. The update of each grid cell is calculated in line~\ref{alg1:line:negative_update} and described in one of the following sections.

Finally, a set of salient locations is calculated in line~\ref{alg1:line:get_salient}. The result of the algorithm is the newly updated map $G$ and a set of salient locations $S$.

\subsection{Positive map update}\label{sec:positive_update}


A vision-based detection consists of a bounding box in an image with a confidence score. Additionally, the camera's pose, which captured the processed image, is considered in the form of its geographical location and orientation. Moreover, general knowledge of the domain is incorporated in the form of an expected real-world size of an object being detected. 

\begin{figure}[t]
    \centering
    \includegraphics[width=1.0\textwidth]{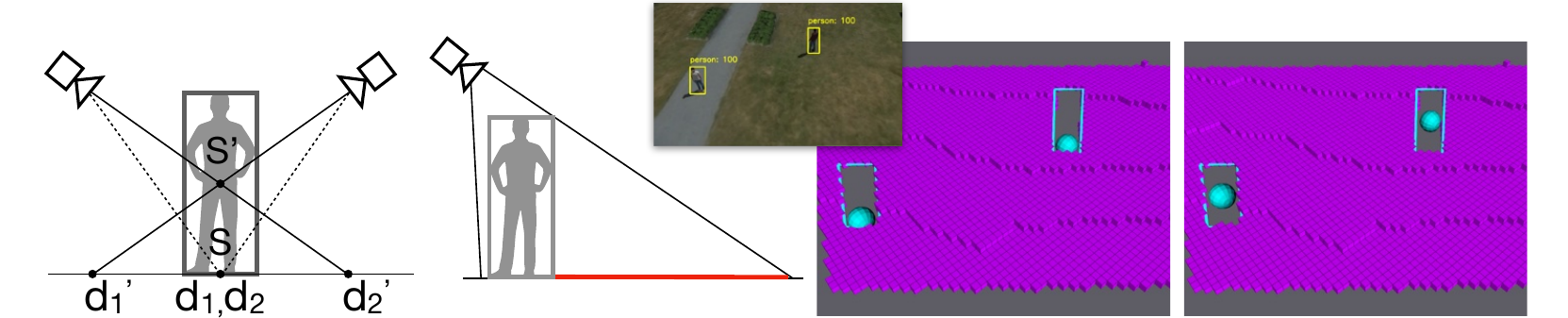}
    \caption{Left: The effect of selecting the significant point of a bounding box when converting to a location in the world. 
    Selecting the center ($S'$) results in different locations ($d1'$ and $d2'$) while the bottom center ($S$) results in a single location ($d1$, $d2$). Middle: Area in red is not updated because it is behind a detected object's bounding box. Right: Update of the grid based on an image frame for bottom and center of a bounding box.}
    \label{fig:bbox_location_hint}
\end{figure}

A location, called a \emph{significant point}, within the bounding box is also specified for a given object class and used during the map update. For certain objects and applications, selecting the center of the bounding box is sufficient. In general, it has a minor influence on the overall accuracy of the estimated salient location. However, it can be crucial when very accurate results are needed. The influence of the point selection is schematically presented in Figure~\ref{fig:bbox_location_hint}.

The positive probability update of a grid cell corresponding to the bounding box's significant point is calculated using several contributing components:

\begin{equation}
P(n\given z_{t})_{positive}=0.5 +  \areafactor \cdot \detfactor \cdot \detscore \cdot \maxpositive
\label{eq:p_positive}
\end{equation}

The base value of the probability is 0.5 (i.e., log-odds $L=0$), which does not change the value of the log-odds/probability (see Equation~\ref{eq:voxelloupdate}). The first two components, $\areafactor \in [0,1]$, and $\detfactor \in [0,1]$, incorporate information related to the bounding box and the relative detector fidelity, respectively. The third, $\detscore \in [0,1]$ is the confidence score of the detection. Finally, the last component $\maxpositive \in [0,0.5]$ allows for scaling, thus limiting the probability update's maximum value. 

The first component, $\areafactor$, considers the information provided by the detection bounding box and domain knowledge, where the expected real-world sizes of detected objects are incorporated. For a given object class and the distance to the grid cell, it is possible to calculate the expected sizes of bounding boxes and quantify whether the obtained result is reasonable. Too big or too small bounding boxes are a result of false positive detections and should, therefore, be trusted less or outright rejected by calculating $\areafactor$ as 0.0.
The area component is calculated taking into account class-specific minimum and maximum object dimensions: $\objareamin, \objareamax$ expressed in meters, the image resolution of a detector network ($\imsize$), and camera intrinsic parameters: focal lengths ($f_{x|y}$) and image size used for camera calibration ($\imsizecalib$). First, an expected range of areas of the bounding boxes for the above-mentioned parameters and the distance between the camera and the object detection ($\camobjdist$) is calculated using a general equation:

\begin{equation}
bbox_{w|h}=\frac{\imsize\cdot f_{x|y}\cdot \objsize}{\imsizecalib\cdot \camobjdist}
\label{eq:bbox_size}
\end{equation}

\begin{figure}[tb]
    \centering
    \includegraphics[width=1.0\textwidth]{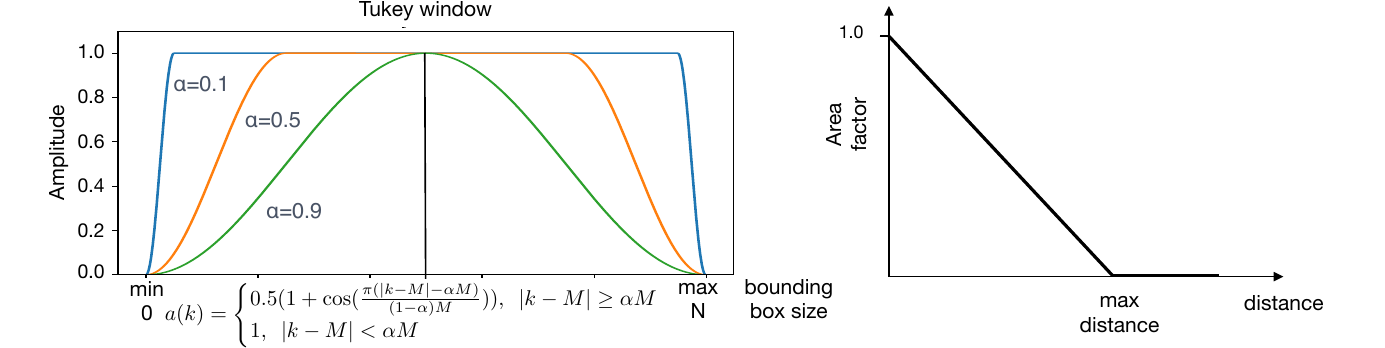}
    \caption{Left: Example shapes of the Tukey window filter for three $\alpha$ values. The bounding box sizes below and above the minimum and maximum expected sizes result in the value of 0 of the component, effectively resulting in outlier rejection. Inside the window (between $min$ and $max$), the influence depends on the given $\alpha$ value. Right: Calculating an area component corresponding to a distance of the observed location. The component is 0 for distances above a maximum distance that corresponds to the minimum object size. It grows linearly for distances smaller than the threshold. }
    \label{fig:tukey}
\end{figure}

Using the equation, the minimum and maximum bounding box sizes can be calculated by substituting object size ($\objsize$) with minimum and maximum $obj_{w|h}^{min|max}$, respectively. Given these quantities, the area component is calculated using a windowing function, for example, the Tukey window~\cite{harris1978use} (also called cosine-tapered) expressed by the equation shown in Figure~\ref{fig:tukey}, which presents example shapes of the filter for three $\alpha$ values. The highest value is achieved for detections in the middle of the size window (i.e., between minimum and maximum). It falls off more or less rapidly depending on the value of $\alpha$. With $\alpha$ values close to 0, it becomes rectangular, and at $\alpha$ values close to 1, it is a bell-shaped window. The intuition  is that the obtained bounding boxes with sizes different from what they should be in reality are given a lower area component value. It allows for rejecting outliers and trusting bounding boxes with expected sizes. 

The second component in Equation~\ref{eq:p_positive}, $\detfactor$, is used to incorporate the information about the relative fidelity of a detection technique used to acquire the result. The relative accuracy measures ($f_c$) for several detectors are obtained by performing comparative evaluations as discussed in Section~\ref{sec:networks}. In general, the value of this component can be calculated by normalizing its accuracy measure $f_c$ over the set of detectors ($\detfactor = f_c / max (D)$).
One way to construct this set is to take into account all potentially available detectors, even if, in the current mission, only a subset is feasible and used. Another approach can be to include only the detectors used for a particular mission. If only a single detector is used, the value of $\detfactor$ is 1 and it does not influence the value of the positive update.

The final component of the update, $\maxpositive$, allows for lowering the influence of a single detection in a map update. This feature is used to adjust the overall behavior of the proposed method, for example,  when detected objects are moving (see Experiment 2 in Section~\ref{sec:experimental}). 

\subsection{Negative map update}\label{sec:negative_update}

The purpose of the negative map update is to incorporate information about the observed locations within the field of view of the camera. When performing this update, the intuition is to quantify \emph{how well a location was observed}, which means how certain it is that the location does not contain objects of specific classes. 
Based on the resulting map, it should then be possible to reason about the need for additional observations or additional offline processing of the collected data using better detectors. 
The computation of the negative update is complementary to the positive update. The former quantifies how well-suited the detector was to make the observation. For the latter, if there was no detection of an object, how certain it is that it is not a false negative.
The negative update is done for all voxels within the camera's field of view, excluding detection bounding boxes (if any) because locations "behind" bounding boxes are not observed (see Figure~\ref{fig:bbox_location_hint}). 

The negative update of a voxel $n$ is given by:
\begin{equation}
P(n\given z_{t})_{negative}=0.5 - \distfactor \cdot \detfactor \cdot \maxnegative
\label{eq:p_negative}
\end{equation}
where, the nominal value equals 0.5. The value of $\distfactor \in [0,1]$ incorporates the information about the distance between the camera and a specific grid cell. The $\detfactor$ is calculated the same way as for the positive update. The $\maxnegative \in [0,0.5]$ component serves the same function as in the case of positive probability updates. Namely, it allows for limiting the influence of a single observation.

\sloppy The distance component $\distfactor$ is obtained by first calculating the $max\_distance$ corresponding to the minimum object size by modifying Equation~\ref{eq:bbox_size} to obtain the distance for a specific object and its expected size. 
Using the obtained value and the distance between the camera and the grid cell ($distance_n$), the distance component is calculated as follows:
\begin{equation}
\distfactor=max(1 - distance_n/max\_distance), 0.0)
\label{eq:dist_factor}
\end{equation}

For distances larger than the maximum, the factor equals 0 and grows linearly to 1 with smaller distances. Intuitively, this factor captures the "range" of the detection ability. The effect of this factor can be seen in Figure~\ref{fig:exp_diff_score} where areas on the edges of the map are less certain as they are further away from the camera. 
A graphical representation of the function is depicted in Figure~\ref{fig:tukey} (Right).

\subsection{Salient locations calculation}\label{sec:extract_salient}

Given an object occupancy map obtained using the method described in the previous sections, a list of salient locations can be extracted. First, a set of grid cell locations with probabilities above a given threshold is obtained (line~\ref{alg1:line:threshold} of Algorithm~\ref{alg:grid_update}). Second, Euclidean cluster extraction is performed on the centers of the obtained grid cells using a distance threshold. The threshold value is selected based on the application at hand, and typical values are 1-3 meters. 

For each cluster of locations obtained, the average probability is calculated as the probability of the salient location. Similarly, the cluster center is calculated by averaging the locations of detections used to update the map. The final value of the location becomes the geographical location of the detected object, the salient point location.

\section{Experimental evaluation}\label{sec:experimental}

\begin{figure}[tb]
    \centering
    \includegraphics[width=1.0\textwidth]{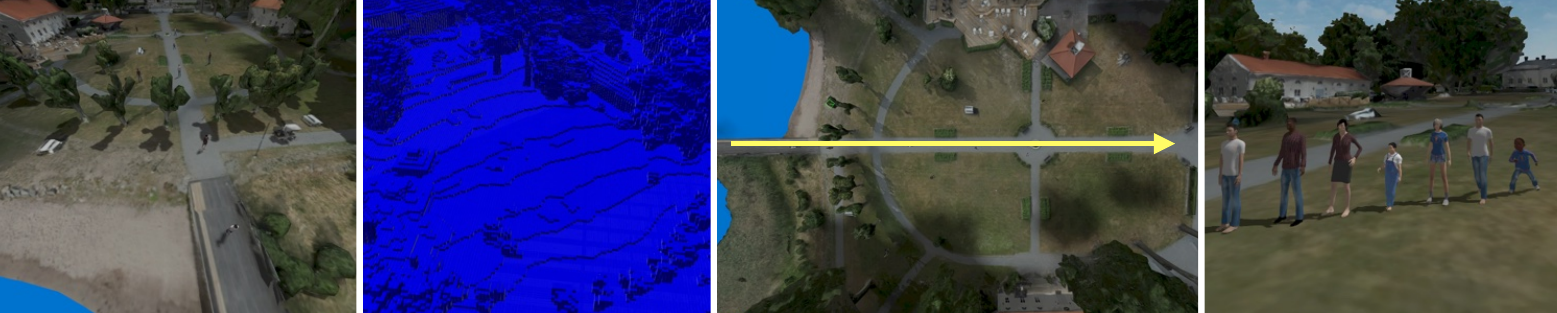}
    \caption{Experimental setup: 1) 2.5D model of the environment, 2) occupancy grid map of built based on the 2.5D model, 3) yellow arrow shows an approximate flight path used in two experiments and 4) 3D models used during simulated experiments.}
    \label{fig:exp_setup}
\end{figure}

Several experiments have been performed in simulation and real flights to validate the proposed approach in practice. 
The left side of Figure~\ref{fig:exp_setup} shows a 2.5D model of the operational environment for both simulated and real experiments. The model has been built using a photogrammetric modeling software\footnote{Agisoft Metashape, https://www.agisoft.com} and over eleven thousand images captured using a DJI Matrice 300 UAV platform and a DJI Zenmuse P1 camera sensor\footnote{https://enterprise.dji.com/matrice-300, https://enterprise.dji.com/zenmuse-p1}. The 2.5D model is used in the ROS2 Gazebo software to simulate a camera sensor to obtain realistic images. Based on the 2.5D model of the operational environment occupancy grid maps have been built with different resolutions: 1m, 0.5m, and 0.25m. An example map with a resolution of 0.5m is presented in the second image of Figure~\ref{fig:exp_setup}.

Both simulated and real experiments use the same API to the UAV. This means that all software is identical for the two kinds of experiments. 
The most computationally intensive components of the presented method are related to obtaining the list of visible voxels (line~\ref{alg1:line:get_indexes} of Algorithm~\ref{alg:grid_update}) as well as computing the distance between the camera and the detected objects. In order to speed up these computations a custom OpenGL
vertex shader is used to create the model based on the existing voxel grid, where a fragment shader is responsible for computing both a depth image as well as the visible cubes image. GPU execution allows calculations to be performed in milliseconds instead of seconds.

\subsection{Experiment 1 - detector accuracy}\label{sec:exp1}

\begin{figure}[tb]
    \centering
    \includegraphics[width=1.0\textwidth]{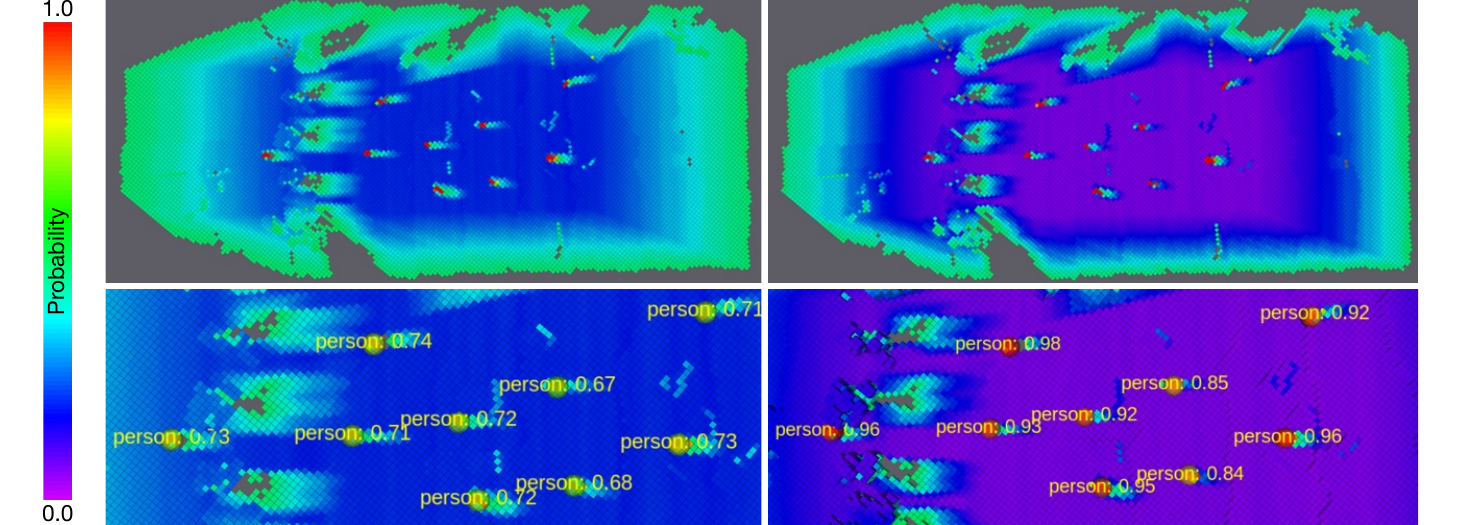}
    \caption{Experiment 1: Influence of the detector relative accuracy score on the probability update. A detector with a lower relative accuracy (left) produces a less certain (blue) map while a more accurate detector produces a more certain (purple) map.}
    \label{fig:exp_diff_score}
\end{figure}

The purpose of this experiment is to demonstrate the influence of the relative accuracy ($\detfactor$ in Equations~\ref{eq:p_positive} and \ref{eq:p_negative}) of the detector used on the resulting map update. The experiment is performed in simulation with eight different 3D models of people (see right of Figure~\ref{fig:exp_setup}) randomly placed at 9 locations in the environment. The UAV is flown, as shown by the yellow arrow on the right side of Figure~\ref{fig:exp_setup}, at an altitude of 15 meters above the ground level and a speed of 3 m/s. The camera is tilted 45 degrees down. The data obtained is processed twice, first with the $\detfactor$ with the nominal value (0.98) and second with the factor artificially cut in half.

By comparing the left and right sides of Figure~\ref{fig:exp_diff_score}, it can be observed that the probability of the salient locations is lower on the left (average of 0.71 vs 0.93) since the lower factor is used. Similarly, the probability of locations not containing objects is higher (less certain). This shows the desired property of the method to differentiate between detectors of different relative accuracies. 

\subsection{Experiment 2 - rate of negative update.}\label{sec:exp_moving_object}

This experiment aims to demonstrate the proposed method's ability to handle non-static objects. In this experiment, a stationary UAV is observing an object moving from left to right within the camera's field of view. The same sequence was processed twice with different values of $\maxnegative$ in Equation~\ref{eq:p_negative}. 

\begin{figure}[tb]
    \centering
    \includegraphics[width=1.0\textwidth]{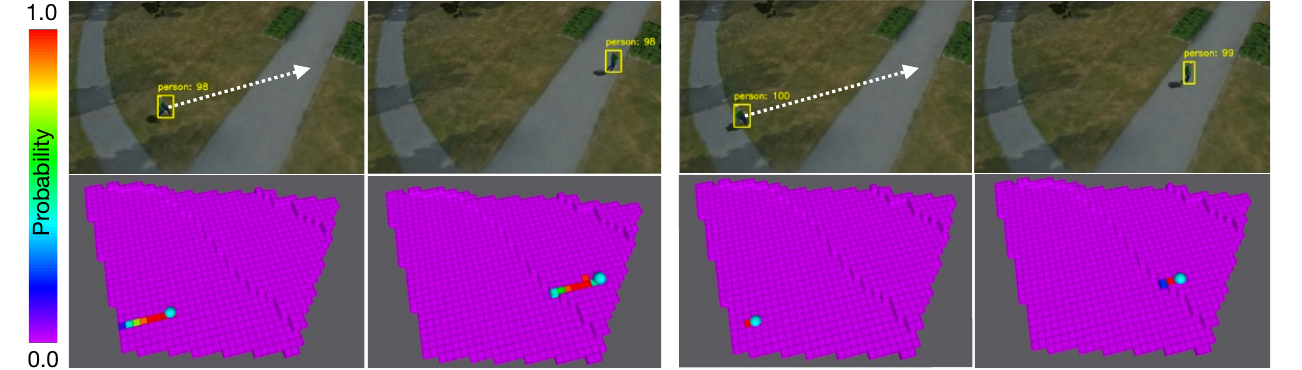}
    \caption{Experiment 2: Influence of the negative update rate. Slow update is visible on the left - a trail behind a moving object is visible. Fast update is visible on the right - the moving object's position is quickly reflected in the map. }
    \label{fig:forgetting}
\end{figure}

The left side of Figure~\ref{fig:forgetting} shows two frames of a video sequence and the corresponding updated probability maps with a low value of $\maxnegative$ of $0.05$. This results in performing the negative probability update to a small degree based on a single frame and relying on the number of frames for the update to be more certain. This behavior is preferred when the image-based detector used is unreliable and does not detect objects in all frames. Negative update would make it be quickly removed from the map. The result of the moving object is a trail of probabilities being slowly updated to reflect that the object has moved.

The right side of Figure~\ref{fig:forgetting} presents the result for $\maxnegative$ of $0.5$. The negative update is performed in a way that incorporates the negative information very quickly. The trail behind a moving object is rapidly removed from the map. 

\subsection{Experiment 3 - influence of the grid size on geolocation accuracy.}

The purpose of this experiment is to show the influence of the used grid resolution on the geolocation accuracy. The experiment is performed with the same setup as in Experiment 1 above. Based on the data collected the method is used three times with different voxel grid resolutions of: 1m, 0.5m, and 0.25m. The ground truth of positions of the detected objects is taken from the Gazebo simulator. 

The average position errors were: 0.63m, 0.36m, 0.27m, respectively for the three grid map resolutions. As expected, the accuracy improves with higher resolution of the grid map but the difference is not significant. 

\subsection{Experiment 4 - geolocation accuracy in a real flight.}

\begin{figure}[tb]
    \centering
    \includegraphics[width=1.0\textwidth]{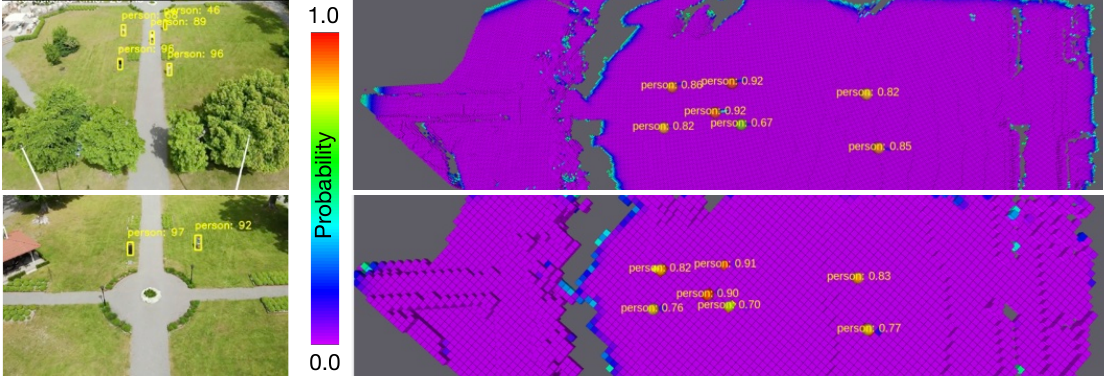}
    \caption{Experiment 4: Real flight experiment results. Two example frames of the live video feeds on the left. Resulting saliency maps (top: 25cm, bottom: 1m) on the right.}
    \label{fig:exp_real}
\end{figure}

The purpose of this experiment is to evaluate the method in a real flight as part of an autonomous UAV mission of Search and Rescue type. The setup of the mission is very similar to the Experiment 1 presented above. Seven volunteers of different ages, heights and genders were positioned in the environment and their locations were  measured using a Real Time Kinematics (RTK) GPS receiver. Three resolutions of the voxel grids were tested: 1m, 0.5m, and 0.25m. The bottom center of the bounding box was used as the significant point (see left of Figure~\ref{fig:bbox_location_hint}) when performing the calculations. Example video frames and maps for two resolutions are presented in Figure~\ref{fig:exp_real}. The resulting average geolocation accuracy for the three resolutions were: 1.15m, 1.02m, and 0.90m, respectively. As expected, the average errors of the geolocation were higher than in case of the simulated experiment presented above. Inaccuracies of the UAV's GPS position (non-RTK) and camera gimbal's, as well as the lack of hardware time synchronisation between the camera and the vehicle all contributed to the increased errors. Nonetheless the overall accuracy was very high in relation to the object sizes.

\subsection{Experiment 5 - detector allocation.}\label{sec:alloc_1}

\begin{figure}[tb]
    \centering
    \includegraphics[width=1.0\textwidth]{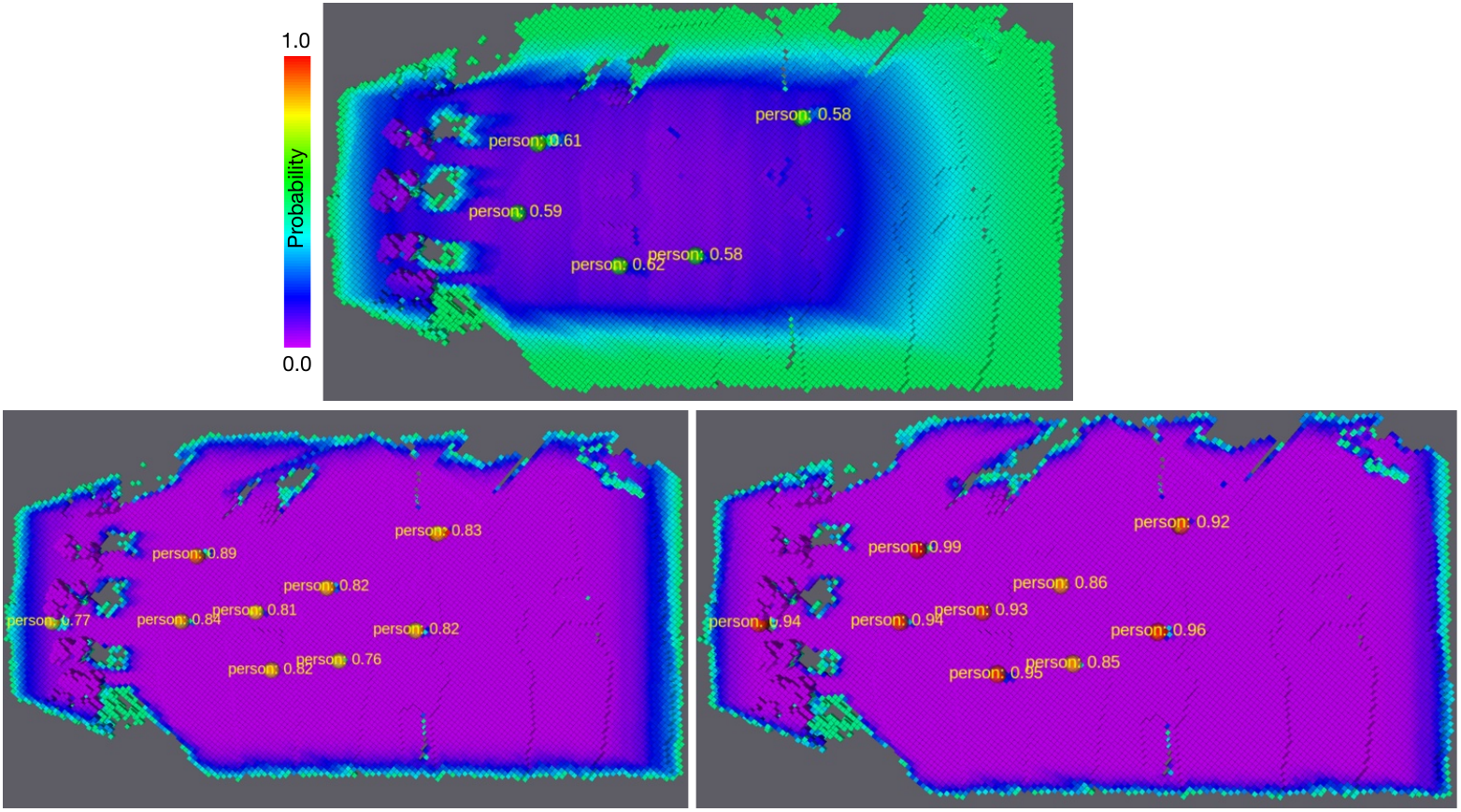}
    \caption{Experiment 5: influence of detector allocation. Top: local robot processing; Bottom left: remote processing at a single site, Bottom right: remote processing at 2 sites.}
    \label{fig:alloc_1}
\end{figure}

The purpose of this experiment is to demonstrate the method for optimally allocating detectors that has been presented in Section~\ref{sec:det_choice}, which takes into account the available computational and communication resources. The experiment is performed with the same setup as in Experiment 1 above. Figure~\ref{fig:alloc_1} presents three maps and salient location results for three different allocations of detectors. The top image depicts a situation where the only processing on a machine of type A (see Table~\ref{tab:hw_configs}) is available and the processing is done using an SSD Inception v2 network on a CPU at 5Hz. In the bottom left image, only one machine of type D is available and the frames are processed with the Faster RCNN Resnet 50 network using a GPU at 10Hz. In the bottom right image, two sites of type D are available and the images are processed using the  Faster R-CNN Inception Resnet v2 LP network at both sites at 5 Hz, which effectively gives 10Hz processing. 

The parameters used for the algorithm presented in Section~\ref{sec:det_choice} were as follows. For the first case, only one site of type A was made available with all algorithms from Table \textbf{~\ref{tab:hw_configs}}. The desired processing rate was 5Hz,  $detPerStream=1$, and $detPerFrame=2$. For the second and third cases, one and two remote sites of type D were made available, respectively. The desired processing rate was 10Hz. For all cases, the H.264 codec was used and the link has a bandwidth of 20Mbps. In the third case, the allocation to two sites meant that the bandwidth was split into two 10Mbps streams. 

As can be observed in the top image of Figure~\ref{fig:alloc_1}, processing of data using a relatively low-accuracy detector results in 5 true positive detections and 4 false negatives. The confidence of detections is relatively low (average of 0.59) and the probabilities in the map where no objects are present are relatively high (low confidence). In contrast, the bottom-left  image shows all 9 objects found and geolocated with higher confidence (average of 0.82) and the confidence for areas of the map not containing objects is also significantly higher. Finally, the bottom-right image in Figure~\ref{fig:alloc_1}, which benefits from using two relatively high-accuracy algorithms, has both higher confidence of detections (average of 0.93) and the confidence of the map not containing objects is, as expected, higher, which can by observed at the edges of the map. 

\section{Conclusion}\label{sec:conclusions}

This research presents a very general method for structuring operational missions, where multi-agent systems of robots (in this case, small-scale UAVs) have the goal of identifying and geolocating objects of different types  using vision-based object detection algorithms in complex operational environments. The method considered has three aspects. In the first, a method for evaluating the use of various object detection methods that takes into account the complexities and constraints associated with outdoor operational missions is proposed. Secondly, an object detector allocation algorithm is proposed to optimally allocate object detection methods to each member of a team of robots based on the evaluation, and prior to a mission. Thirdly, A novel method for probabilistic fusion of detection results, geolocation of objects, and generation of saliency maps of identified objects is specified and tested. These resulting saliency maps can then be utilized by rescuers or other robotic systems to collaboritively plan missions, such as delivering food or medical supplies.

The proposed methods have been tested in both simulated and real-world scenarios using teams of small-scale UAVs to demonstrate the versatility of the solutions. The accuracy of the geolocation of objects has been verified using a high-accuracy RTK measurement system.

\subsubsection*{Acknowledgments}
This work has been supported by the ELLIIT Network Organization for Information and Communication Technology, Sweden (Project B09), the Wallenberg AI, Autonomous Systems and Software Program (WASP) funded by the Knut and Alice Wallenberg Foundation, and Sweden's Innovation Agency Vinnova (Projects: 2022-00086, 2023-01035, 2024-01322). The 2nd and 4th authors are also supported by a research grant from Mahasarakham University, Thailand.

\bibliographystyle{apalike}
\bibliography{paper}



\end{document}